\documentclass[10pt,journal,compsoc]{IEEEtran}

\usepackage{newtxtext, newtxmath}
\usepackage{graphicx}
\usepackage{amsmath}
\usepackage{amssymb}
\usepackage{xspace}
\usepackage{xcolor}
\usepackage{comment}
\usepackage{wrapfig}
\usepackage{float}
\usepackage{fancyhdr}
\usepackage{lastpage}
\usepackage{color}
\usepackage{contour}
\usepackage{xkeyval}
\usepackage{multirow}
\usepackage{color}
\definecolor{darkblue}{rgb}{0.0,0.0,0.6}
\newcommand{\cog}{COG\xspace}

\usepackage{bm}

\ifCLASSOPTIONcompsoc
  \usepackage[nocompress]{cite}
\else
  \usepackage{cite}
\fi

\ifCLASSINFOpdf
\else
\fi

\begin{document}
\title{Clouds of Oriented Gradients for 3D Detection of Objects, Surfaces, and Indoor Scene Layouts}

\author{Zhile~Ren
        and~Erik~B.~Sudderth%
\IEEEcompsocitemizethanks{\IEEEcompsocthanksitem Z.~Ren is with School of Interactive Computing, Georgia Institute of Technology, Atlanta, GA, 30332.\protect \hfil\break
E-mail: jrenzhile@gmail.com
\IEEEcompsocthanksitem E.~B.~Sudderth is with the School of Information and Computer Sciences, University of California, Irvine, CA, 92697-3435.\protect \hfil\break
E-mail: sudderth@uci.edu}%
}

\IEEEtitleabstractindextext{%
\begin{abstract}
We develop new representations and algorithms for three-dimensional (3D) object detection and spatial layout prediction in cluttered indoor scenes. We first propose a clouds of oriented gradient (COG) descriptor that links the 2D appearance and 3D pose of object categories, and thus accurately models how perspective projection affects perceived image gradients. To better represent the 3D visual styles of large objects and provide contextual cues to improve the detection of small objects, we introduce 
latent support surfaces. We then propose a ``Manhattan voxel'' representation which better captures the 3D room layout geometry of common indoor environments. Effective classification rules are learned via a latent structured prediction framework. Contextual relationships among categories and layout are captured via a cascade of classifiers, leading to holistic scene hypotheses that exceed the state-of-the-art on the SUN RGB-D database.
\end{abstract}

\begin{IEEEkeywords}
3D scene understanding, object detection, room layout estimation, structured prediction, cascaded classification.
\end{IEEEkeywords}}

\maketitle

\IEEEdisplaynontitleabstractindextext

\IEEEpeerreviewmaketitle

\newcommand{\etal}{\emph{et al}.}
\newcommand\todo[1]{\textcolor{red}{#1}}

\IEEEraisesectionheading{\section{Introduction}\label{sec:introduction}}
\IEEEPARstart{S}{emantic} understanding of three-dimensional (3D) scenes plays an increasingly important role in modern robotic systems and autonomous vehicles. The last decade 
has seen major advances in semantic understanding of 2D images~\cite{pascal-voc-2012,ILSVRC15}.
However, images of indoor (home or office) environments remain challenging for existing methods due to the prevalence of clutter and occlusions.
Advances in depth sensor technology can reduce ambiguities in standard RGB images, enabling breakthroughs in scene layout prediction~\cite{lee2009geometric, hedau2010thinking,zhang2013estimating}, support surface prediction~\cite{silberman2012indoor,fouhey2014unfolding, guo2013support}, 
semantic parsing~\cite{gupta2014learning}, and object detection~\cite{song2014sliding,song2016ssc,ren2016cog}. A growing number of annotated RGB-D datasets have been constructed to train and evaluate indoor scene understanding methods~\cite{russell2009,lai2011,silberman2012indoor,song2015sun}. 

Holistic indoor scene understanding~\cite{song2015sun} requires integrated detection of objects and the room layouts (walls, floors, and ceilings) that surround them.
While object detection is often formalized as the prediction of a 2D bounding box~\cite{pascal-voc-2012}, 2D representations are insufficient for many real-world applications because they do not explicitly represent object orientations or contextual relationships. %
We instead propose to detect the 3D size, position, and orientation of object instances via bounding \emph{cuboids} (convex polyhedra). 
3D cuboid detection is a standard task in indoor and outdoor scene understanding benchmarks~\cite{song2015sun,Geiger2012kitti}. 

Descriptors constructed from point cloud representations of RGB-D images are frequently used for 3D object detection.  For example, Song~\etal~\cite{song2016deep} use the \emph{truncated signed distance function} (TSDF) to define descriptors for candidate 3D bounding cuboids.
But given the diverse variation in the appearance of indoor object categories,
accurately modeling how appearance varies with object style and 3D viewpoint is
very challenging~\cite{fidler20123d}. 
We thus design a novel, orientation-adaptive gradient descriptor that uses perspective geometry to better detect objects observed from diverse 3D viewpoints.

Basic discriminative scene parsing algorithms detect each category independently, but often have many false positives.
Previous work has used manually engineered heuristics to prune false detections~\cite{song2014sliding} or combined
CAD models with layout cues to model scenes~\cite{Geiger2015GCPR}. 
In this paper, we significantly boost detection accuracy via a \emph{cascaded classification framework}~\cite{heitz2009cascaded} that learns contextual relationships among object categories, 
as well as relationships between objects and the overall room layout.
This efficient approach allows initial detections of visually distinctive objects to lead to holistic scene interpretations of higher quality.
 
To estimate the spatial layout used in our cascaded classifiers, we assume an orthogonal ``Manhattan'' room structure~\cite{coughlan1999manhattan}.   
Many previous methods predict 2D projections of the underlying 3D room structure~\cite{schwing2012efficient,bai2012fast}, but small 2D alignment errors may lead to poor 3D layout estimates.
We avoid this by using a structured prediction framework to directly estimate 3D layouts from RGB-D images, and propose a \emph{Manhattan voxel} representation that (like our object descriptors) is adapted to the geometry of indoor scenes.
Our learning-based approach is more robust to the noisy depth estimates produced by practical RGB-D cameras, and thus avoids errors made by simpler layout prediction heuristics~\cite{song2015sun}.

Holistic indoor scene understanding is particularly challenging because smaller objects, like lamps and monitors, only occupy a tiny fraction of the room volume.  
Bottom-up detectors thus have high computational demands (many candidate bounding cuboids must be considered) and typically produce many false positives.
To address this challenge, we note that many small objects are \emph{supported} by the surfaces of large objects~\cite{silberman2012indoor,guo2013support}, and augment our cuboid representations with latent support surfaces.  While surface heights are estimated without explicit training annotations, modeling them nevertheless boosts the accuracy of our furniture detectors.  When integrated into our cascaded classification framework, support surfaces constrain the search space for small objects, and thereby improve detection speed as well as accuracy.

In summary, we propose a general framework for learning detectors for multiple object categories using only RGB-D annotations. 
We first introduce a \emph{cloud of oriented gradient} (\cog) descriptor that robustly links 3D object pose to 2D image boundaries, and discuss extensions that further boost performance (Sec.~\ref{sec:cog}).
Because a major cause of feature inconsistency across object instances is variation in the location of the supporting surface, we model this height as a latent variable, and use it to distinguish different visual styles and detect smaller objects (Sec.~\ref{sec:latent}).  
We also introduce a \emph{Manhattan voxel} representation to predict room layout directly from RGB-D data (Sec.~\ref{sec:layout}).
We use a structured prediction framework to learn an algorithm that aligns 3D cuboid hypotheses to RGB-D data, and a \emph{cascaded classifier} to incorporate contextual cues from other object instances and categories, as well as the overall 3D layout (Sec.~\ref{sec:cascade}).
We evaluate our algorithm on the challenging SUN RGB-D dataset~\cite{song2015sun} and achieve state-of-the-art accuracy in the 3D detection of 19 object categories (Sec.~\ref{sec:experiments}).

\section{Related Work}
\label{sec:related}

Two-dimensional object detection is a widely studied problem.
Dalal and Triggs~\cite{dalal2005histograms} introduced the~\emph{histogram of oriented gradient} (HOG) descriptor to model 2D object appearance using image gradients. Building on HOG, Felzenszwalb~\etal~\cite{felzenszwalb2010dpm} used a discriminately-trained part-based model to represent objects. This method is effective because it explicitly models object parts as latent variables and thus captures some object style and pose variations. 
More recently, many papers have used \emph{convolutional neural networks} (CNNs) to extract rich features from images~\cite{girshick2014rich,girshick15fastrcnn,ren2015faster,he2016res,lin2017focal}. For domains where large sets of labeled images are available, CNNs lead to state-of-the-art performance with efficient detection speed~\cite{redmon2016you,redmon2016yolo9000}.

Increasingly, real-world computer vision systems often incorporate depth data as an additional input to increase accuracy and robustness. With depth maps we can reconstruct point cloud representations of scenes, leading to significant recent advances in
3D object classification~\cite{wu20143d,su15mvcnn}, 
point cloud segmentation~\cite{qi2017pointnet,qi2017pointnetpp}, 
cuboid-based geometric modeling~\cite{hao2013,jia2013,xiao2012},
room layout prediction~\cite{lee2017roomnet,schwing2013box}, 
3D contextual modeling~\cite{shao2014imagining,zhang2017context},
and 3D shape reconstruction~\cite{pamishapeTulsianiKCM15,dai2017scannet}. 
Here, we focus on the related problem of 3D object detection.

In outdoor scenes, localizing objects with 3D cuboids has become a standard in the popular KITTI autonomous 
driving benchmark~\cite{Geiger2012kitti}. 3D detection systems model car shape and occlusion
patterns using LiDAR or stereo inputs~\cite{chen173dop,mousavian20173d,xiang2015data,qi2017frustum,zhou2017voxelnet}, 
and may also incorporate additional overhead imagery~\cite{chen17multiview3D}. 3D cuboid representations
are more powerful than 2D bounding boxes because they contain more information about 3D object locations, physical
occupancy, and orientation. However, 
many outdoor 3D detection systems are specialized to vehicles and pedestrians, 
and may not generalize to cluttered indoor environments.

Other work has localized indoor objects with 3D cuboids~\cite{lin2013holistic,gupta2010blocks}, but achieving high accuracy is challenging due to the significant shape variations found in cluttered, real-world environments.
Several recent methods have incorporated CAD models to learn object shape~\cite{wu20143d,guptaCVPR15align,song2014sliding} or hallucinate alternative viewpoints for appearance-based matching~\cite{Aubry14chair,lim2013parsing,lim2014fpm}.
While CAD models are a potentially powerful information source, there does not exist an 
abundant supply of models for all categories, and many methods are limited to a small number of object categories~\cite{Aubry14chair}.
Moreover, example-based methods~\cite{song2014sliding} may be computationally inefficient
due to the need to match each exemplar to each image.

For robotics applications, a 3D convolutional neural network was designed to detect simple objects in real time~\cite{maturana2015voxnet}.
In 2015, Song~\etal~introduced a SUN RGB-D dataset~\cite{song2015sun} containing 
10,335 RGB-D images with accurate 3D cuboid annotations for indoor objects, room layouts, and scene categories. The size of the dataset matches
that of the PASCAL-VOC dataset~\cite{pascal-voc-2012} and motivates several recent research projects.
Some methods utilize pre-trained 2D detectors and region proposals as priors~\cite{ILSVRC15}, and
localize 3D bounding boxes via a separate CNN~\cite{song2016deep,zhuo17amodal3d,lahoud20172d,qi2017frustum}. These
methods are efficient and can achieve decent accuracy, but are sensitive to failures of the 2D object detector, which may not generalize to objects seen from novel 3D viewpoints.

Detecting support surfaces is an essential first step in understanding the geometry of 3D scenes for such tasks as
surface normal estimation~\cite{wang2015designing,fouhey2014unfolding} and shape retrieval~\cite{bansal2016marr}.
Silberman~\etal~\cite{silberman2012indoor} use semantic 
segmentation to model object support relationships; this work was
later extended by Guo~\etal~\cite{guo2013support} for support surface prediction. 
We instead use 3D support surface representations to improve the accuracy of our models of object style, and the speed of our detectors for small object categories.

To enable more holistic understanding of 3D scenes, we also predict the locations of walls, ceilings, and floors; this structure is sometimes called the \emph{room layout}~\cite{song2015sun}.
Some related work has predicted 2D projections of the 3D layout~\cite{schwing2013box,zhang2013estimating,schwing2012efficient,hedau2009recovering,lee2017roomnet,mallya2015learning}, or used CNNs to directly predict the 3D layout~\cite{zou2018layout}. 
In this paper, we use the geometric structure of typical indoor environments to design a \emph{Manhattan voxel} representation that leads to accurate 3D layout predictions.

More broadly, holistic scene understanding systems integrate forms of semantic object reasoning, spatial context modeling, and scene type identification~\cite{yao2012describing, song2015sun}. Often, models for each sub-task are learned independently, and then integrated via \emph{conditional random fields} (CRFs) like that proposed by Lin~\etal~\cite{lin2013holistic}.  However, rich scene models lead to complex graph structures and challenging inference problems. Hoiem~\etal~\cite{hoiem2008putting} jointly estimate the camera viewpoint and detect objects, Zhang~\etal~\cite{zhang2017context} use pre-defined room configurations to adjust object localizations, while Ren~\etal\cite{yuzhuo2018C3D} utilize scene type to refine detector outputs. We instead adapt the \emph{cascaded prediction} framework~\cite{heitz2009cascaded} to learn multi-stage models capturing detector accuracies and contextual relationships among objects and the room layout.

\begin{figure*}[!ht]
  \centering
  \includegraphics[width=0.9\linewidth]{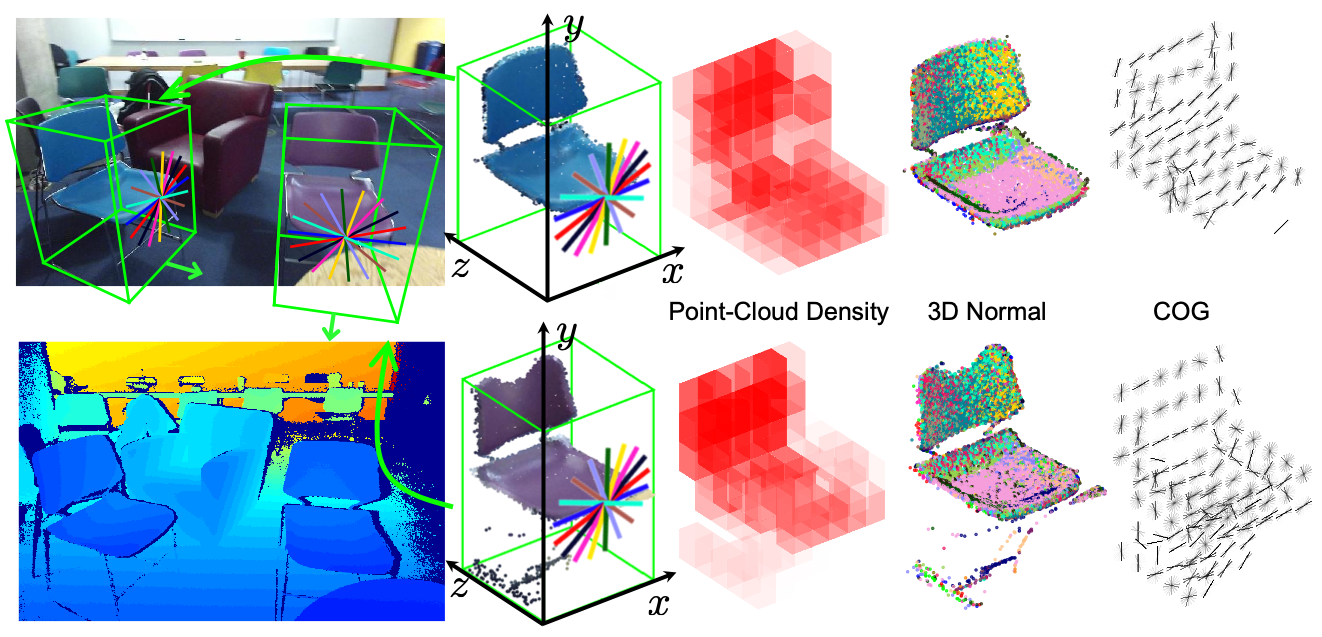}
  \caption{Given input RGB and Depth images (left), we align oriented cuboids and transform observed data into a canonical coordinate frame.  
  For each voxel, we then extract (from left to right) point cloud density features, 
  3D normal orientation histograms, and \cog descriptors of back-projected image gradient orientations.
  \cog bins (left) are colored to show the alignment between instances. 
The value of the point cloud density feature is proportional to the voxel intensity, 
each 3D orientation histogram bin is assigned a distinct color,
and \cog features are proportional to the normalized energy in
each orientation bin, similarly to HOG descriptors~\cite{dalal2005histograms}.}
\label{fig:cog_bins}
\end{figure*}

\section{Modeling 3D Geometry \& Appearance}
\label{sec:cog}

Feature extraction is one of the most important steps for object detection algorithms. 2D object detectors typically
use either hand-crafted features based on image gradients~\cite{dalal2005histograms,felzenszwalb2010dpm} or learned features
from deep neural networks~\cite{girshick2014rich,girshick15fastrcnn,ren2015faster,he2016res,lin2017focal}. For 3D object 
detection systems with additional depth inputs, Gupta~\etal~\cite{gupta2014learning} use horizontal disparity, height above the ground, 
and the angle of the local surface normal to encode images as a three channel (HHA) map for learning with CNNs.  While convolutional processing of 2D images may be used to extract features from 2D bounding boxes, it does not directly model 3D cuboids.
Song~\etal~propose a deep sliding shape~\cite{song2016deep} method that 
combines TSDF features~\cite{song2014sliding} with standard
2D CNN features to describe 3D cuboids, but do not explicitly model 3D cuboid orientation.

Our object detectors are learned from 3D oriented cuboid annotations in the SUN-RGBD dataset~\cite{song2015sun}.
We discretize each cuboid into a $5\times5\times5$ grid of (large) voxels, and extract features for these $5^3=125$ cells.  Voxel dimensions are scaled to match the size of each instance. We use standard descriptors for the 3D geometry of the observed depth image, and propose a novel \emph{cloud of oriented gradient} (\cog) descriptor of RGB appearance. We also introduce simple extensions that improve its performance.

\subsection{Object Geometry: 3D Density and Orientation}
\subsubsection{Point Cloud Density}
Conditioned on a 3D cuboid annotation or detection hypothesis $i$, suppose voxel $\ell$ contains $N_{i\ell}$ points.  We use perspective projection to find the silhouette of each voxel in the image, and compute the area $A_{i\ell}$ of that convex region.  The \emph{point cloud density} feature for voxel $\ell$ then equals $\phi^a_{i\ell} = N_{i\ell}/A_{i\ell}$.  Normalization gives robustness to depth variation of the object in the scene.  We normalize by the local voxel area, rather than by the total number of points in the cuboid as in some related work~\cite{song2014sliding}, to give greater robustness to partial object occlusions.

\subsubsection{3D Normal Orientations}
Various representations, such as spin images~\cite{johnson1999using}, have been proposed for the vectors normal to a 3D surface.  As in~\cite{song2014sliding}, we build a 25-bin histogram of normal orientations within each voxel, and estimate the normal orientation for each 3D point via a plane fit to its 15 nearest neighbors. This feature $\phi^b_i$ captures the surface shape of cuboid $i$ via patterns of local 3D orientations.

\subsection{Clouds of Oriented Gradients (\cog)}

The \emph{histogram of oriented gradient} (HOG) descriptor~\cite{dalal2005histograms} forms the basis for many effective object detection methods~\cite{pascal-voc-2012}.  Edges are a very natural foundation for indoor scene understanding, due to the strong occluding contours generated by common objects.  However, as gradient orientations are determined by 3D object orientation and perspective projection, HOG descriptors that are naively extracted in 2D image coordinates generalize poorly.

\begin{figure}[t]
\centering
   \includegraphics[width=0.99\linewidth]{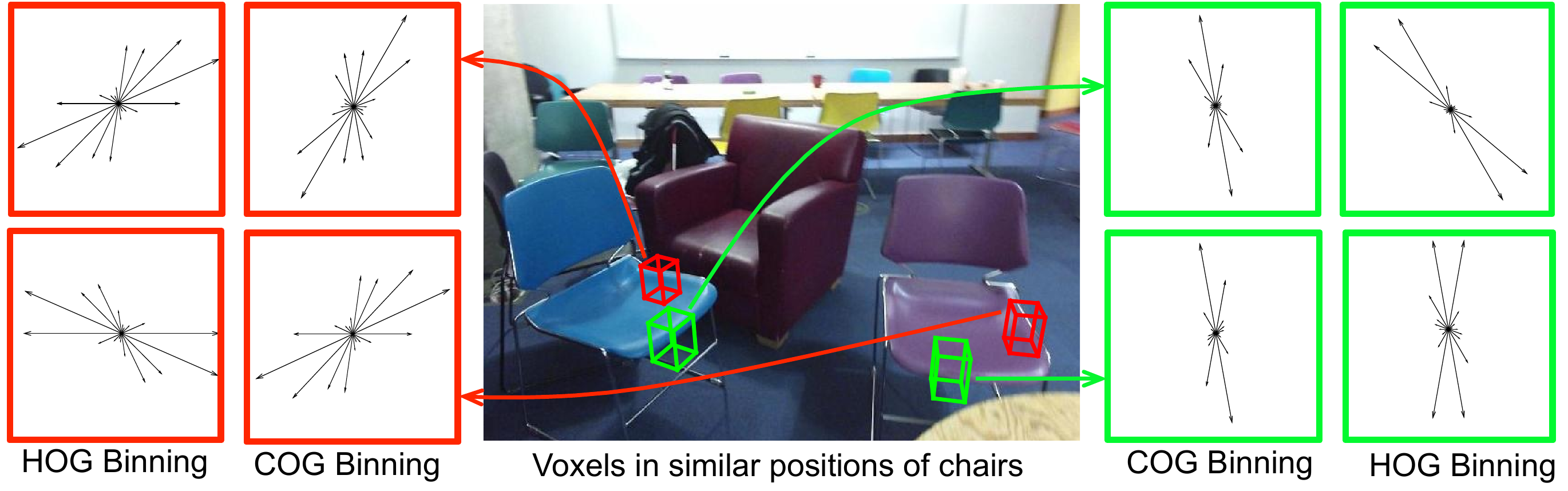}
 \caption{For two corresponding voxels (red and green) on two chairs, we illustrate the orientation histograms that would be computed by a standard HOG descriptor~\cite{dalal2005histograms} in 2D image coordinates, and our \cog descriptor in which perspective geometry is used to align descriptor bins.  Even though these object instances are very similar, their 3D pose leads to wildly different HOG descriptors.}
\label{fig:cog_hog}
\end{figure}

To address this issue, some previous work has restrictively assumed that parts of objects are near-planar so that image warping may be used for alignment~\cite{fidler20123d}, or that all objects have a 3D pose aligned with the global ``Manhattan world coordinates'' of the room~\cite{hedau2010thinking}. 
The \emph{bag of boundaries} (BOB)~\cite{payet2011contours} descriptor builds separate gradient-based models for each of several distinct 3D viewpoints, rather than using geometry to generalize across 3D viewpoints.
Some previous 3D extensions of the HOG descriptor~\cite{buch20093d, scherer2010histograms} assume that a full 3D model is given. 
In recent work~\cite{song2016}, 3D cuboid hypotheses were used to aggregate standard 2D features from a deep convolutional neural network, but the deep features are not conditioned on object orientations.
Our \emph{cloud of oriented gradient} (\cog) feature accurately describes the 3D appearance of objects with complex 3D geometry, as captured by RGB-D cameras from any viewpoint.

\subsubsection{2D Gradient Computation} We compute gradients by applying filters $[-1, 0, 1]$, $[-1, 0, 1]^T$ to the RGB channels of the unsmoothed 2D image. The maximum responses across color channels are the gradients $(dx,dy)$ in the $x$ and $y$ directions, with corresponding magnitude $\sqrt{dx^2 + dy^2}$.
We follow similar implementation details to the gradient computations used in HOG descriptors~\cite{dalal2005histograms}. 
The 2D unsigned gradients are then aggregated in each voxel to define our 3D \cog descriptor.

\subsubsection{3D Orientation Bins} 
The standard HOG descriptor~\cite{dalal2005histograms} for cell~$\ell$ of object~$i$ uses nine evenly spaced gradient histogram bins, 
$(\bm{o}_{i\ell}^{(1)}, \dots, \bm{o}_{i\ell}^{(9)})$.
For all object instances,
$\bm{o}_{i\ell}^{(1)} = [1, 0]^T$ is aligned with the horizontal image direction.
As shown in Fig.~\ref{fig:cog_hog}, HOG descriptors may thus be inconsistent for (even nearly identical) objects in distinct poses. 

Because objects from the same category typically have similar local 3D structure, for each oriented 3D cuboid proposal, we instead model local gradient statistics in a canonical 3D coordinate frame.
As illustrated in Fig.~\ref{fig:cog_bins}, we define nine evenly spaced 3D orientation bins 
$(\bm{O}_{i\ell}^{(1)}, \dots, \bm{O}_{i\ell}^{(9)})$
on the front surface ($xy$-plane) of each voxel~$\ell$ within the cuboid.
For all instances, $\bm{O}_{i\ell}^{(1)}$ %
is aligned with the horizontal 3D $x$-axis
(dark blue lines in Fig.~\ref{fig:cog_bins}). 
Given the camera's intrinsic matrix $K$, and the extrinsic matrix $[R_i|t_i]$
encoding the relative 3D pose of cuboid~$i$,
we use perspective projection to map 3D orientation bins to 2D image coordinates:
\begin{equation}
\begin{bmatrix}
  \bm{o}_{i\ell}^{(j)}\\ 1 
\end{bmatrix}
\propto K\,[R_i|t_i]
\begin{bmatrix}
  \bm{O}_{i\ell}^{(j)}\\ 1
\end{bmatrix}\!.
\end{equation}
This transform aligns the 2D orientation bins for distinct 3D cuboids.
For each pixel that back-projects to 3D voxel~$\ell$, we accumulate its unsigned 2D gradient in the corresponding projected orientation bin to define a nine-dimensional \cog feature $\phi^c_{i\ell}$.

Some previous work has warped images to align with fixed 2D orientation 
bins~\cite{hedau2010thinking}, but such affine transformations may be unstable for
objects with non-planar geometry. Our \cog descriptor can be seen as accumulating standard gradients with warped histogram bins, rather than warping images to match fixed orientation bins.  This innovation enables our later learning algorithms to better generalize to novel 3D views of complex objects.

\subsubsection{Normalization and Aliasing}  We bilinearly interpolate gradient magnitudes between neighboring orientation bins~\cite{dalal2005histograms}.  To normalize the histogram $\phi^c_{i\ell}$ for voxel $\ell$ in cuboid $i$, we then set $\phi^c_{i\ell} \leftarrow \phi^c_{i\ell} /\sqrt{||\phi^c_{i\ell}||^2+\epsilon}$ for a small $\epsilon > 0$. 
Accounting for all orientations and voxels, the dimension of the \cog feature $\phi_i^c$ is $5^3\times9=1125$.

\subsection{Extensions of the COG Descriptor}

\subsubsection{View-to-Camera Features}
\begin{figure}[!ht]
\centering
\includegraphics[width=0.9\linewidth]{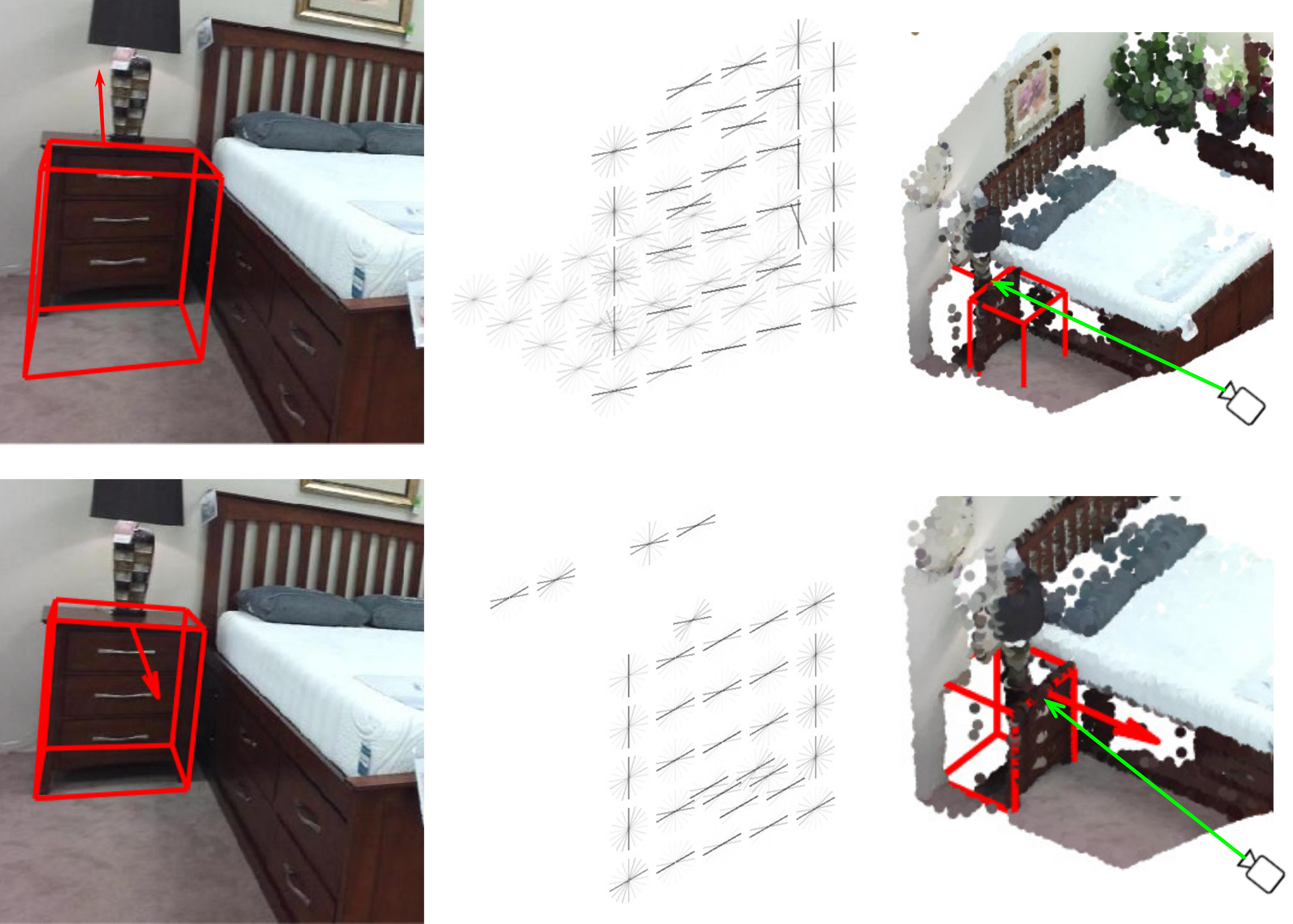}
 \caption{A false positive 3D detection for the nightstand category that occurs without a view-to-camera feature  (top). 
          The COG feature is similar to that of a correct detection (bottom) due to bilateral symmetry.}
   \label{fig:exp_view}
\end{figure}

For single view RGB-D inputs, objects like nightstands and other furniture may only expose one planer surface to the camera. 
At test time, the features of a 3D cuboid proposal oriented away from the camera may
resemble those of a correct detection (see Fig.~\ref{fig:exp_view})
because voxel features are computed by first rotating the cuboid to a canonical 
coordinate frame. However, due to the self-occlusions that occur in real objects, the features modeled by the COG descriptor would in fact not be visible when objects are facing away from the camera.   
Therefore, we add features to represent 
objects' orientation with respect to the camera, and learn to distinguish implausible object hypotheses.

Specifically, we compute the cosine $x$ of the angle between the cuboid orientation 
and its viewing angle from camera in horizontal direction. Then we define a set of radial basis functions of the form
\begin{equation}
 f_j(x) = \exp\bigg(-\frac{(x-\mu_j)^2}{2\sigma^2}\bigg),
 \label{eq:rbf}
\end{equation}
and space the basis function centers $\mu_j$ evenly between $[-1, 1]$ with step size 0.2.  The bandwidth $\sigma=0.5$ was chosen using validation data.
Radial basis expansions are a standard non-linear regression method, 
and can be seen as a layer of a neural network. 
We expand the camera angle using this basis representation plus a bias feature, producing an 11-dimensional \emph{view-to-camera} feature $\phi^d_{i}$.

\subsubsection{Expanded Cuboid Features}
Many object detection systems have a pre-processing stage that generates bounding box proposals that contain objects with well-defined boundaries, instead of amorphous background areas~\cite{alexe2012measuring}. Using a region proposal network to maximize the ``objectness'' score of predicted bounding boxes~\cite{kuo2015deepbox} is thus an essential first step for many state-of-the-art object detection systems~\cite{girshick2014rich,song2016deep}. 

Objectness scores are usually determined from the difference between local and surrounding appearances of each object. 
Instead of designing a separate pre-processing step, we build such contextual cues into our cuboid features. For each cuboid proposal, we expand its size to capture an additional layer of voxels in each direction, so that each cuboid is now described by $7\times7\times7$ voxels. 

\begin{figure}[b]
\centering
\includegraphics[width=0.95\linewidth]{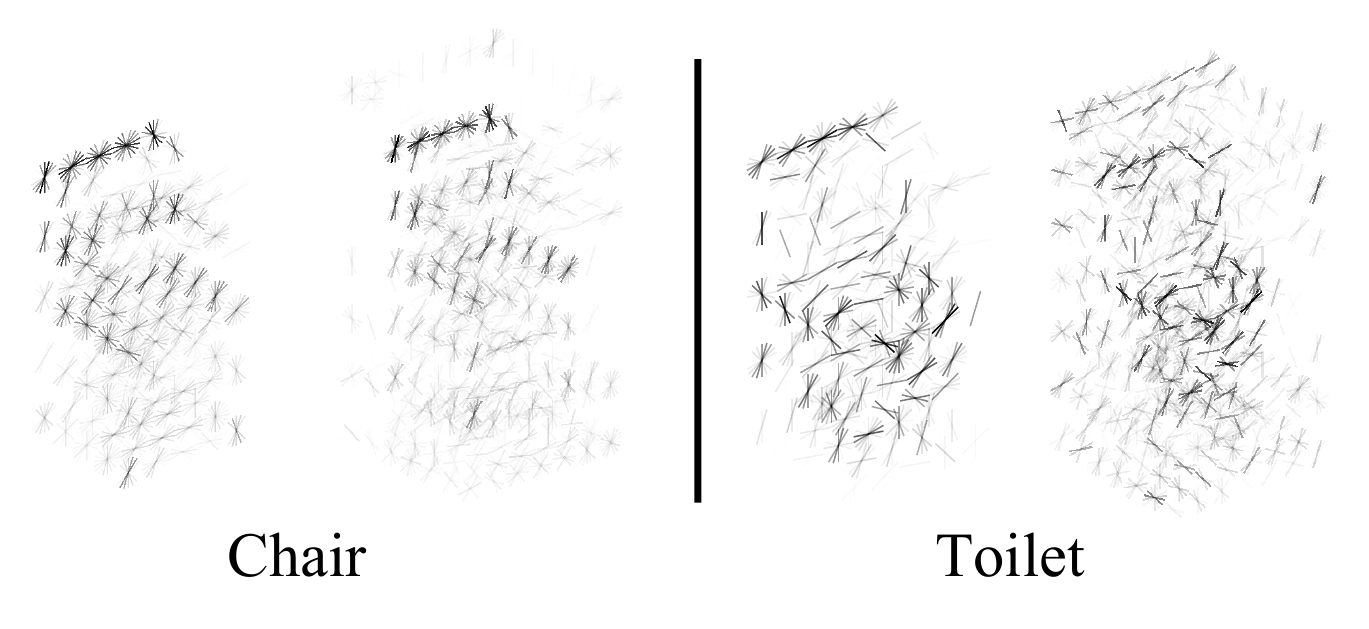}
  \caption{Visualizing the learned weights for the COG (left) and expanded COG (right) features. Although chairs and toilets have similar geometric structures, the appearance of the 3D environment immediately surrounding them is different, producing local contextual cues captured by our expanded COG features.}
   \label{fig:expanded_cog}
\end{figure}

Before discussing the training algorithm, we preview the learned weights of \cog descriptors for the chair and toilet categories in Fig.~\ref{fig:expanded_cog}. Toilets are typically placed against the wall in cluttered bathrooms, while there is typically free space around chairs, and thus our expanded cuboid features capture differences between these categories that improve detection accuracy.

The structure of our expanded cuboid feature has some similarities to the ``zoom-out'' features originally proposed for 2D image segmentation~\cite{mostajabi2015feedforward}, and used by Song~\etal~\cite{song2016deep} for 3D detection. We provide ablation studies in Table~\ref{table:ablation}, and demonstrate that this extension is very effective in modeling the geometric structure surrounding each cuboid, improving object detection accuracy.

\subsection{Structured Prediction of Object Cuboids}
\label{sec:ssvm}
For each voxel $\ell$ in some cuboid $B_i$ annotated in training image $I_i$, we have one point cloud density feature $\phi^a_{i\ell}$, 25 surface normal histogram features $\phi^b_{i\ell}$, and 9 COG appearance features $\phi^c_{i\ell}$. For each cuboid~$i$, we have 12 camera view features $\phi^d_{i}$. Using expanded features with $7^3=343$ voxels, our overall representation of cuboid~$i$ is then
$\phi(I_i,B_i) = [\{\phi^a_{i\ell},\phi^b_{i\ell},\phi^c_{i\ell}\}_{\ell=1}^{343}, \phi^d_{i}]$. 
Cuboids are aligned via annotated orientations as illustrated in Fig.~\ref{fig:cog_bins},
using the gravity direction provided in the SUN-RGBD dataset~\cite{song2015sun}. 

For each object category $c$ independently, using those images which contain visible instances of that category, our goal is to learn a prediction function 
$h_c: I \rightarrow B$ that maps an RGB-D image $I$ to a 3D bounding box
$B = (L, \theta, S, y)$.  Here $L$ is the center of the cuboid in 3D, $\theta$ is the cuboid orientation, $S$ is the physical size of the cuboid along the three axes determined by its orientation, and $y$ is a binary variable indicating whether the object is present in that area of the 3D scene. We assume objects have a base upon which they are typically supported, and thus $\theta$ is a scalar rotation with respect to the ground plane. %

Given $n$ training examples of category $c$, we use an $n$-slack formulation of the structural support vector machine (SVM) objective~\cite{joachims2009cutting} with margin rescaling constraints:
\begin{equation}
\begin{split}
\min_{w_c, \xi\geq 0}\quad & \frac{1}{2}w_c^Tw_c + \frac{C}{n}\sum_{i=1}^n\xi_i \\
\text{subject to} \quad
&w_c^T[\phi(I_i, B_i) - \phi(I_i, \bar{B}_i)] \geq \Delta(B_i, \bar{B}_i) - \xi_i, \\
\text{for all}\quad& \bar{B}_i \in \mathcal{B}_i, i = 1,\ldots,n. 
\label{eq:svm}
\end{split}
\end{equation}
Here $\phi(I_i,B_i)$ are the features for oriented cuboid hypothesis $B_i$ given RGB-D image $I_i$, $B_i$ is the ground-truth cuboid annotation, and $\mathcal{B}_i$ is the set of possible alternative cuboids.
For training images with multiple instances, as in previous work on 2D detection~\cite{vedaldi09structured} we add multiple copies to the training set, each time removing the subset of 3D points contained in other instances.

Given some ground truth cuboid $B$ and estimated cuboid $\bar{B}$, we define the loss function as follows. If a scene contains ground truth cuboid B and indicator variable $\bar{y}=1$, we compute
\begin{equation}
  \Delta(B, \bar{B}) = 1 - \text{IOU}(B, \bar{B})\cdot
  \left(\frac{1 + \cos(\bar{\theta}-\theta)}{2}\right).
  \label{eq:loss}
\end{equation}
Here, $\text{IOU}(B,\bar{B})$ is the volume of the 3D intersection of the cuboids, divided by the volume of their 3D union.  The loss is bounded between 0 and 1, and is smallest when the $\text{IOU}(B,\bar{B})$ is near 1 \emph{and} the orientation error $\theta-\bar{\theta} \approx 0$.  The loss approaches~1 if either position or orientation is completely wrong. If a scene does not contain any ground truth 
instances of the object and 
the indicator variable $\bar{y}=0$ for the cuboid proposal, the loss equals~0. We penalize all other cases with a loss of 1. We solve the loss-sensitive objective of Eq.~\eqref{eq:svm} using a cutting-plane method~\cite{joachims2009cutting}.

\subsection{Cuboid Hypotheses}
We create cuboid proposals in a sliding-window fashion using discretized
3D world coordinates, with 16 candidate orientations. We
discretize cuboid sizes using empirical statistics of the cuboid annotations
in the training database:
$\{0.1, 0.3, 0.5, 0.7, 0.9\}$ width quantiles, 
$\{0.25, 0.5, 0.75\}$ depth quantiles, and $\{0.3, 0.5, 0.8\}$
height quantiles. Every combination of cuboid size, 3D
position on the ground plane (whose height is estimated as described in Sec.~\ref{sec:layout}), and 3D orientation is then evaluated.

\subsection{Relative Importance of 3D Cuboid Features}
We explore the relative importance of different features for the detection of 5 large objects
in Table~\ref{table:ablation}. We first trained our detector with geometric 
features only (Geom), with \cog only (COG), with both geometric and \cog features (Geom+\cog), 
adding the camera-view feature (Geom+COG+view), and finally utilizing the expanded 
cuboid feature (Geom+COG+view+expanded). 
The \cog feature and geometric features
have complementary advantages in 3D object detection, and combining
them leads to improved performance. 
The average accuracies of object detectors improve when additional features are added, 
demonstrating that each step of our feature design is effective.

\begin{table}[t]
\centering
\begin{tabular}{c|c|c|c|c|c} 
 &\includegraphics[width=.03\textwidth]{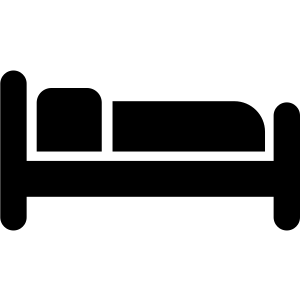}&\includegraphics[width=.03\textwidth]{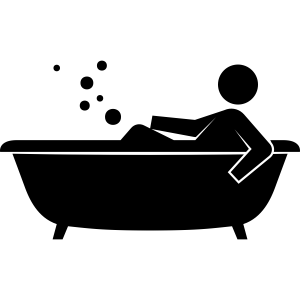}&\includegraphics[width=.03\textwidth]{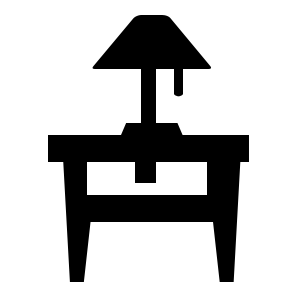}&\includegraphics[width=.03\textwidth]{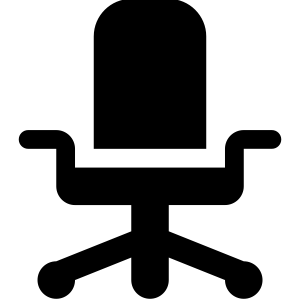}&\includegraphics[width=.03\textwidth]{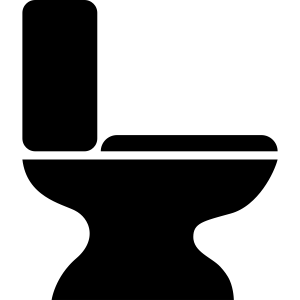} \\ \hline  
\scriptsize Geom&45.0&37.9&2.3&36.2&55.2 \\  
\scriptsize COG&52.5&42.8&6.7&22.6&49.6 \\
\scriptsize Geom+\cog     &53.0&49.8&12.8&39.0&63.6 \\ 
\scriptsize Geom+COG+view     &52.8&53.2&16.8&40.4&57.8 \\ 
\scriptsize  Geom+COG+view+expanded     &63.8&63.8&29.2&64.1&80.5  
\end{tabular}
  \caption{Average precision scores for five object categories (\emph{bed, bathtub, nightstand, chair, toilet}) given various sets of 3D cuboid features.}
\label{table:ablation}
\end{table}

\section{Modeling Latent Support Surfaces}
\label{sec:latent}

\begin{figure}[b]
\centering
\includegraphics[width=0.8\linewidth]{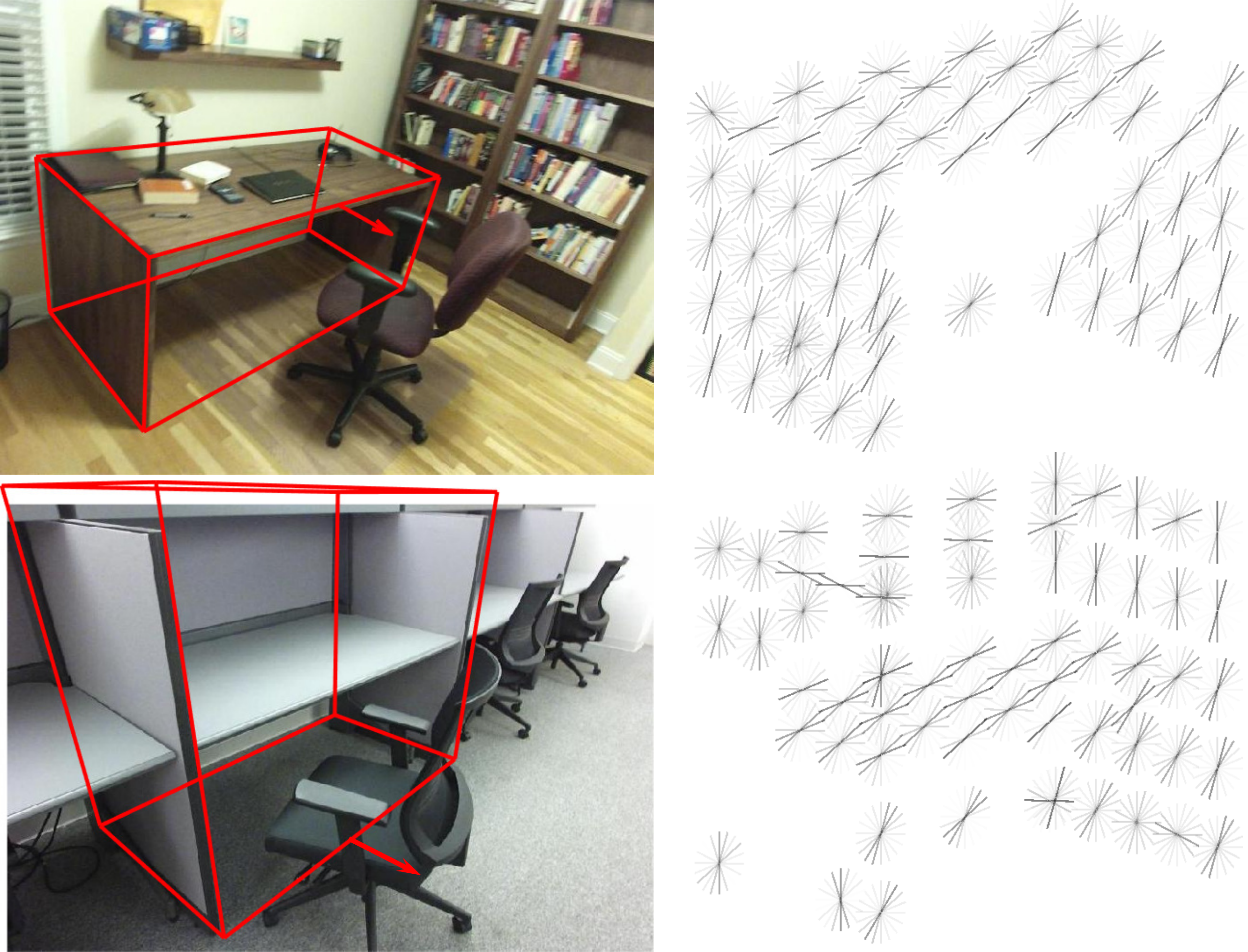}
 \caption{Different surface heights for instances of the ``desk'' category in SUN the
  RGB-D dataset~\cite{song2015sun} lead to inconsistent 3D \cog representations.}
   \label{fig:exp_inconsistent}
\end{figure}

Geometric descriptors and COG descriptors are able to capture local shapes and appearances, 
but objects have widely varying visual styles.
Moreover, 3D cuboids are labeled by different 
annotators from Mechanical Turk to construct the SUN RGB-D dataset~\cite{song2015sun},  and thus objects in the same category
may have inconsistent 3D annotations. As a result, voxel features are sometimes noisy 
and inconsistent across different object instances (see Fig.~\ref{fig:exp_inconsistent}).

To explicitly model different visual styles within each object category, a classical approach is to 
use part-based models~\cite{felzenszwalb2010dpm,fidler20123d}
where objects are explained by spatially arranged parts. 
For many object categories, the height of the support surface 
is the primary cause of style variations (Fig.~\ref{fig:exp_inconsistent}).
Therefore, we explicitly model the support surface as a latent part for each object.

By modeling support surfaces we can also constrain the search 
space for small object detectors. Such detectors are otherwise  
computationally challenging to learn, and perform poorly due to the large set of 3D pose hypotheses.

\begin{figure*}[t]
\centering
\includegraphics[width=0.98\linewidth]{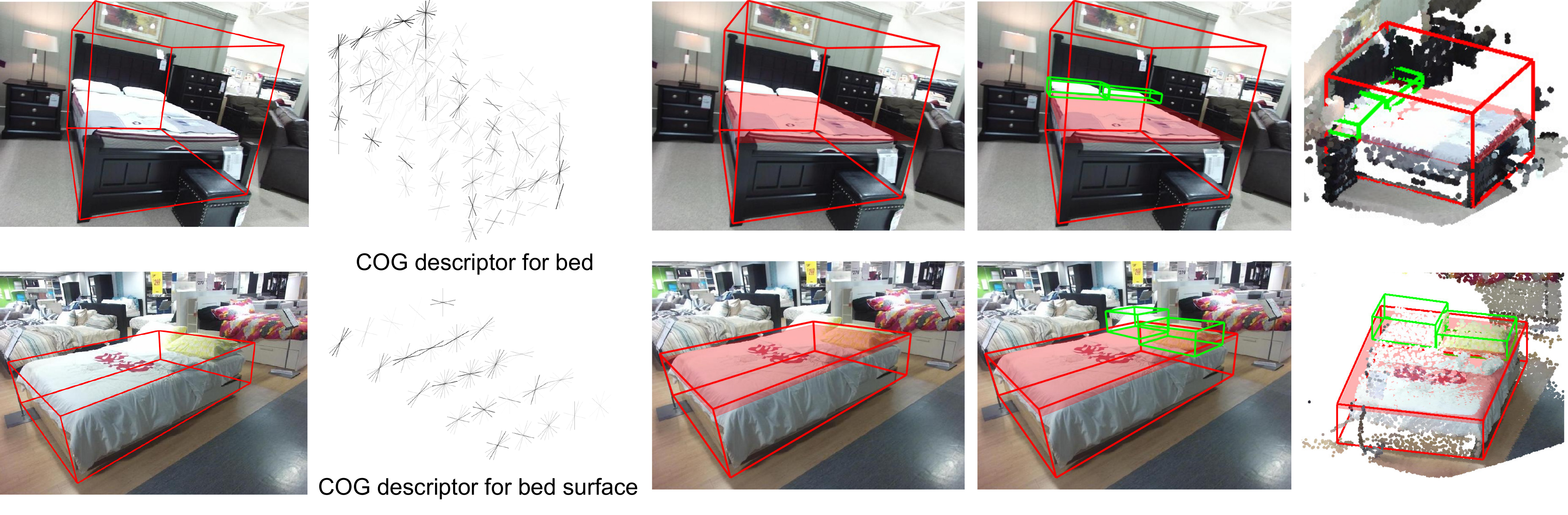}
\caption{Visualization of 3D detection of beds and pillows using latent support surfaces. Given input RGB-D images,  we use our learned COG descriptor to localize 3D objects and infer latent support surfaces (shaded) for 3D proposals of beds ({red}). Then we search for pillows ({green}) that lie on top of the inferred support surfaces. }
   \label{fig:teaser_latent}
\end{figure*}

\subsection{Latent Structural SVM Learning}
Some previous work was specifically designed to predict support surface regions~\cite{guo2013support} from labeled training data,
but the predicted support surfaces are not semantically meaningful. We instead 
treat the height of the support surface of each object as a latent 
variable and use latent structural SVMs~\cite{yu2009learning,felzenszwalb2010dpm} to learn the detector. 

\begin{figure}[t]
\centering
\includegraphics[width=0.7\linewidth]{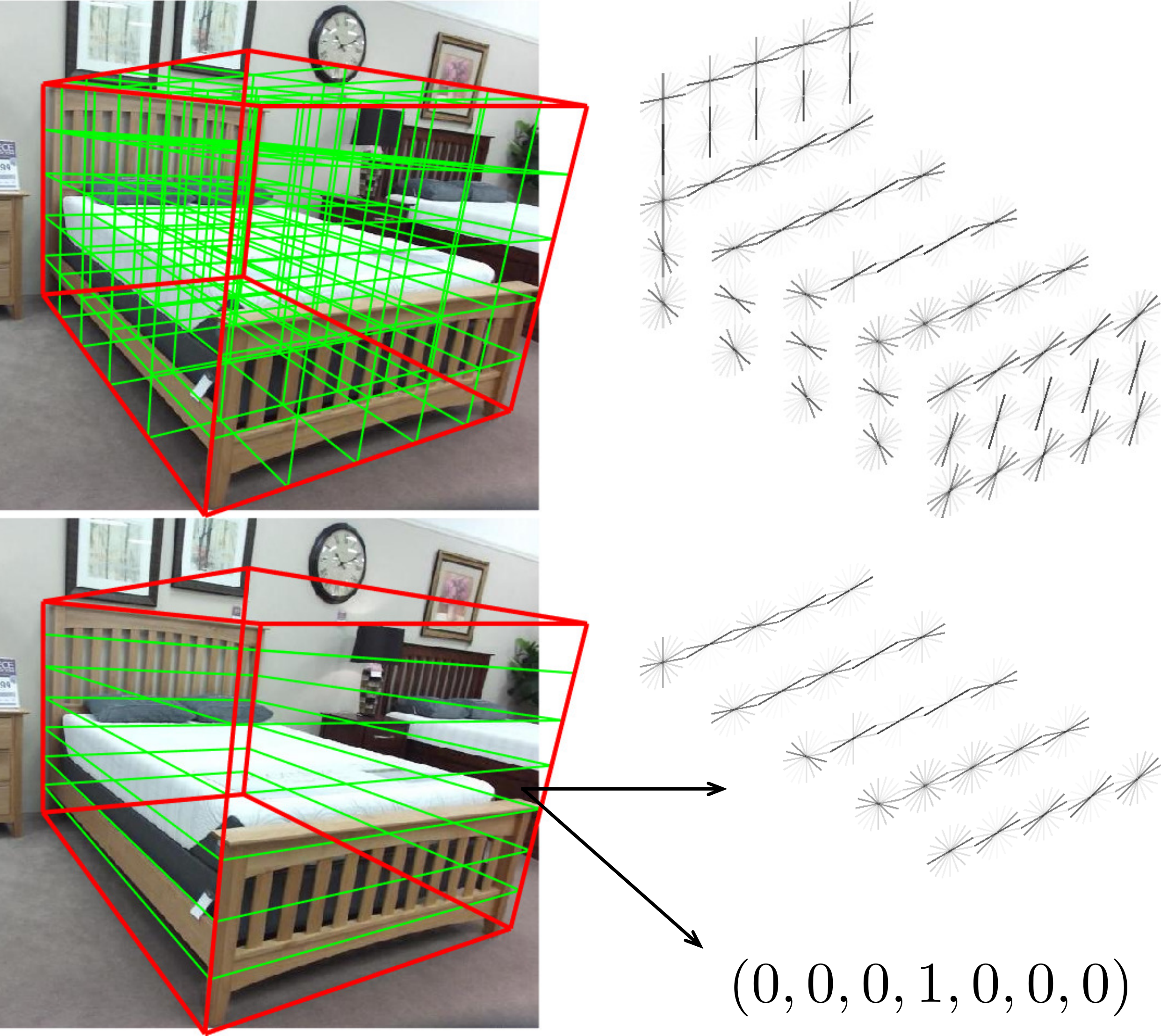}
 \caption{\cog features for 3D cuboids and support surfaces. The surface feature is computed within a single slice of the cuboid, and concatenated with an indicator vector encoding the relative height. Expanded cuboid features are not visualized.}
   \label{fig:latent_feature}
\end{figure}

We follow the notation in Sec.~\ref{sec:ssvm} with an updated objective. 
For each category $c$, our goal is to learn a prediction function $I \rightarrow (B, h)$ that maps an RGB-D image $I$ to a 3D 
bounding box $B = (L, \theta, S, y)$ along with its relative surface height $h$. 
The latent variable $h$ is defined as the relative surface height to the bottom of the cuboid. We
discretize cuboid height to 7 slices, and thus $h$ localizes the support surface to one of those slices (see Fig.~\ref{fig:latent_feature}).

Given $n$ training examples of category $c$, we want to solve the following optimization problem:
\begin{equation*}
\begin{aligned}
&\min_{w_c, \xi\geq 0}\quad \frac{1}{2}w_c^Tw_c + \frac{C}{n}\sum_{i=1}^n\xi_i 
\quad\text{subject to}\\
&\max_{h_i\in\mathcal{H}} w_c^T \phi(I_i, B_i, h_i) - \max_{\bar{h}_i\in\mathcal{H}} w_c^T\phi(I_i, \bar{B}_i, \bar{h}_i) \\
& \geq \Delta(B_i, \bar{B}_i, \bar{h}_i) - \xi_i, \ \text{ for all } \bar{B}_i \in \mathcal{B}_i,\; i = 1,\ldots,n. 
\end{aligned}
\end{equation*}
Here $B_i$ is the target cuboid, $\mathcal{B}_i$ is the set of possible 
cuboids, and $\mathcal{H}$ is
the set of possible surface heights. $\phi(I, B, h)$ are the features associated to cuboid 
$B$ whose relative surface height is indicated by $h$. We first discretize $B$ into $5\times5\times5$ voxels and 
compute geometric, COG, view-to-camera, and expanded cuboid features, as denoted by $\phi_{\text{cuboid}}(I, B)$.
Then we discretize $B$ with finer resolutions at the vertical dimension into $5\times5\times7$ voxels
and take the $h$-th slice from the bottom to represent cuboid feature, as denoted by $\phi_{\text{surface}}(I, B, h)$. Finally
we add an indicator vector for support surface height, so that
\begin{equation*}
\phi(I, B, h) = [\phi_{\text{cuboid}}(I, B), \phi_{\text{surface}}(I, B, h), 0, ..., 1, ..., 0].
\end{equation*}
We use the same loss function defined in Sec.~\ref{sec:ssvm}.

To train the model with latent support surfaces, we first pre-train cuboid descriptors 
(geometric features, COG, view-to-camera, and scene layout features) without modeling support surfaces.
We then extract the center slice of pre-trained cuboid descriptors and concatenate it to the pre-trained models. Finally, we
initialize the support surface height indicator vector randomly in $[0, 1]$. With this informative initialization, we find that the CCCP algorithm~\cite{yuille2003concave} is effective at solving the (non-convex) 
latent structural SVM learning problem~\cite{yu2009learning}.

\subsection{Small Object Detection via Supporting Surfaces}
While indoor scenes typically contain some large furniture like beds and chairs, 
many other objects with comparatively small physical size are very challenging to detect~\cite{song2016deep,ren2016cog}. 
Some algorithms are specifically designed to detect small objects in 2D images
using multi-scale methods~\cite{chen2016r,hu2016finding}, 
but they cannot be directly applied to 3D object detection.

A severe issue for detecting small objects is that the search space can be enormous, and
thus training and testing with sliding-window cuboid proposals can be computationally intractable. 
But note that small objects, 
such as pillows and monitors and lamps, are usually placed on top of other objects
with support surfaces. If we only search for small objects on predicted support surfaces,
the search space will be greatly reduced. As a result, the inference speed will be improved 
and object proposals will have fewer false positives. This is another benefit of modeling 
support surfaces.

In our implementation, we first detect large objects and furniture that rest 
on the ground.  Then using the cascaded detection framework described in Sec.~\ref{sec:cascade}, we only search for smaller objects on top of
the support surfaces of those large objects with positive confidence scores. 
We reduce the voxel grid to $3\times3\times3$ for lamps and pillows due to their small size, and to $3\times1\times3$ for monitors and TVs due to their flat shape.

\section{Room Layout Geometry: Manhattan Voxels}
\label{sec:layout}

Given an RGB-D image, indoor scene parsing requires not only object detection, but also room layout (floor, ceiling, wall) prediction~\cite{hedau2009recovering, lee2009geometric, zhang2013estimating, schwing2013box}.  
Such ``free space'' understanding is crucial for applications like robot navigation.  
Simple RGB-D layout prediction methods~\cite{song2015sun} work by fitting planes to the observed point cloud data, but are sensitive to outliers. We propose a more accurate learning-based approach to predicting Manhattan geometries that utilizes our \cog descriptor.

\begin{figure}[t]
\centering
   \includegraphics[width=0.85\linewidth]{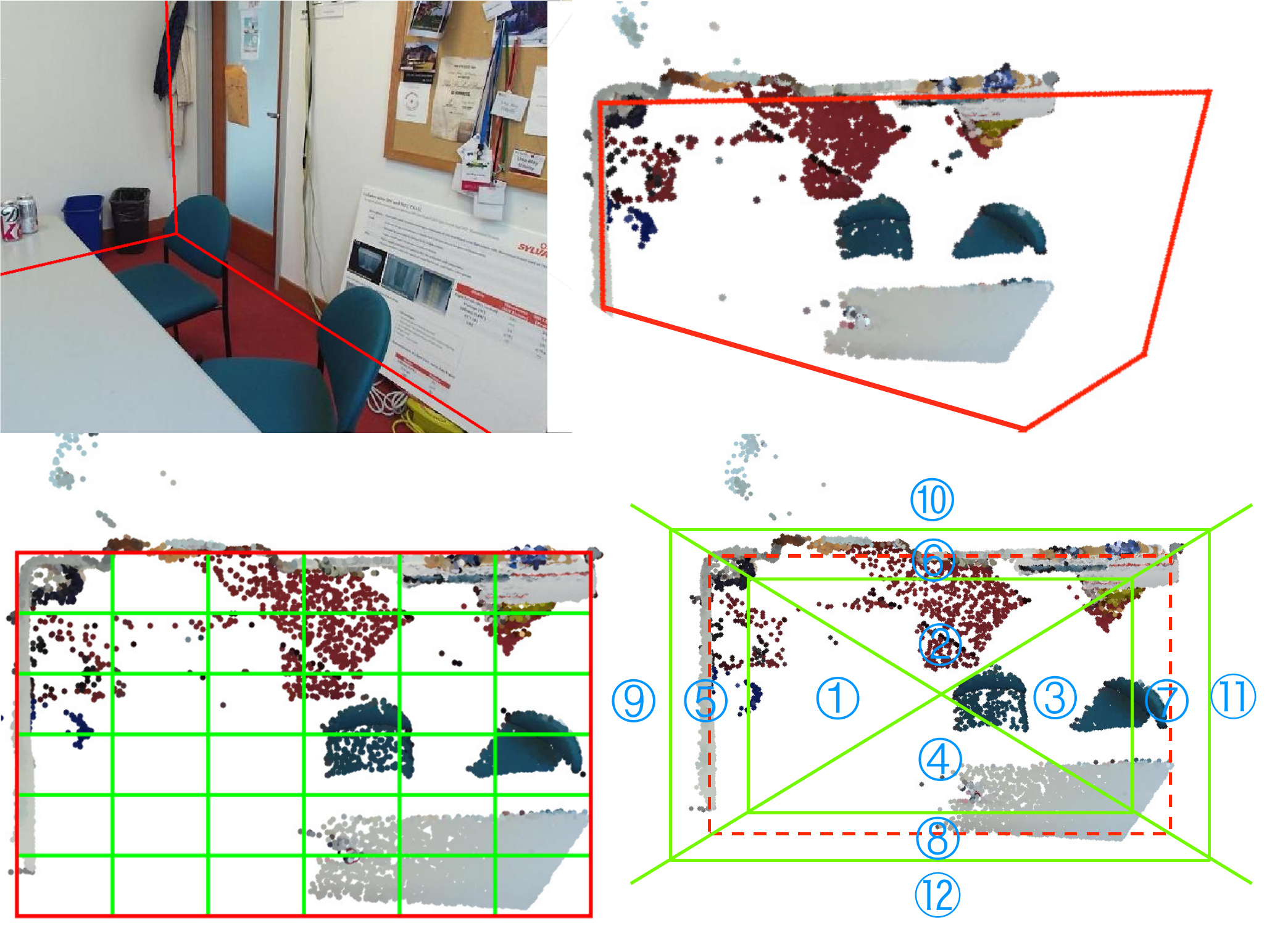}
   \caption{Models for the 3D layout geometry of indoor scenes. \emph{Top:} Ground truth annotation. \emph{Bottom:} Top-down view of the scene and two voxel-based quantizations. We compare a regular voxel grid (left) to our Manhattan voxels (right; dashed red line is the layout hypothesis).}
\label{fig:layout_exp}
\end{figure}

The orthogonal walls of a standard room can be represented via a cuboid~\cite{pero2011sampling}, and we could define geometric features via a standard voxel discretization (Fig.~\ref{fig:layout_exp}, bottom left). 
However, because corner voxels usually contain the intersection of two walls, they then mix 3D normal vectors with very different orientations.  This discretization also ignores points outside of the hypothesized cuboid, and may match subsets of rooms with wall-like structure.

We propose a novel \emph{Manhattan voxel} (Fig.~\ref{fig:layout_exp}, bottom right) discretization for 3D layout prediction.  We first discretize the vertical space between floor and ceiling into 6 equal bins.  We then use a threshold of $0.15m$ to separate points near the walls from those in the interior or exterior of the hypothesized layout.  Further using diagonal lines to split bins at the room corners, the overall space is discretized in $12 \times 6 = 72$ bins.
For each vertical layer, regions $R_{1:4}$ model the scene interior whose point cloud distribution varies widely across images. Regions $R_{5:8}$ model points near the assumed Manhattan wall structure: $R_5$ and $R_6$ should contain orthogonal planes, while $R_5$ and $R_7$ should contain parallel planes. Regions $R_{9:12}$ capture points outside of the predicted layout, as might be produced by depth sensor errors on transparent surfaces.

We again use the S-SVM formulation of Eq.~\eqref{eq:svm} to predict Manhattan layout cuboids $M = (L, \theta, S)$.  
The loss function $\Delta(M, \bar{M})$ is as in Eq.~\eqref{eq:loss}, except we use the ``free-space'' IOU defined by~\cite{song2015sun}, and account for the fact that orientation is only identifiable modulo $90^\circ$ rotations.
Because layout annotations do not necessarily have Manhattan structure, the ground truth layout is defined as the cuboid hypothesis with the largest free-space IOU.

We predict floors and ceilings as the 0.001 and 0.999 quantiles of the 3D points along the gravity direction, and discretize orientation into 18 evenly spaced angles between $0$ and $180^\circ$. We then propose layout candidates that capture at least $80\%$ of all 3D points, and are bounded by the farthest and closest 3D points. For typical scenes, there are 5,000 to 20,000 layout hypotheses.

\begin{figure*}[!ht]
\centering
\includegraphics[width=0.3\linewidth]{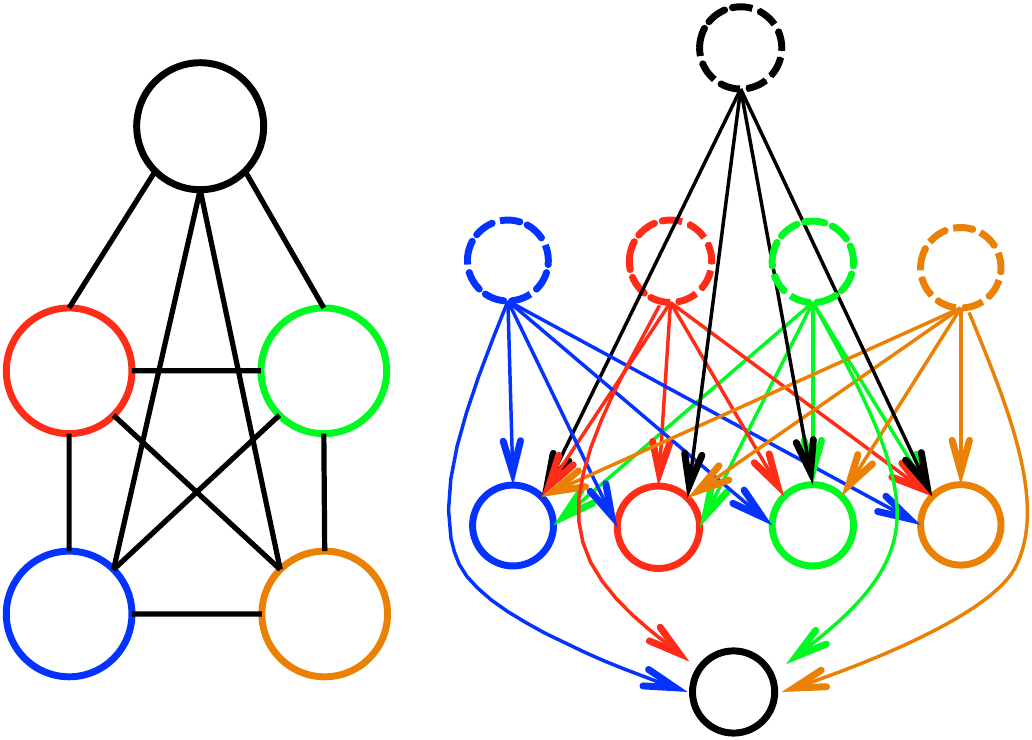}
   \includegraphics[width=0.65\linewidth]{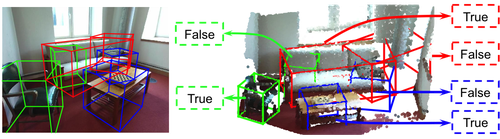}
 \caption{Cascaded classifiers capture contextual relationships among objects. From left to right: (i) A traditional undirected MRF representation of contextual relationships.  Colored nodes represent object categories, and black nodes represent the room layout. (ii) A directed graphical representation of cascaded classification, where the first-stage detectors are hidden variables (dashed) that model contextual relationships among object and layout hypotheses (solid).  Marginalizing the hidden nodes recovers the undirected MRF. (iii) First-stage detections independently computed for each category as in Sec.~\ref{sec:ssvm}. (iv) Second-stage detections (Sec.~\ref{sec:cascade}) efficiently computed using our directed representation of context, and capturing contextual relationships between objects and the overall scene layout.}
   \label{fig:exp_cascade}
\end{figure*}

\section{Cascaded Learning of Spatial Context}
\label{sec:cascade}
If the learned object detectors are independently applied for each category, there may be many false positives where a ``piece'' of a large object is detected as a smaller object (see Fig.~\ref{fig:exp_cascade}).  Song et al.~\cite{song2014sliding} reduce such errors via a heuristic reduction in confidence scores for small detections on large image segments.  To avoid such manual engineering, which must be tuned to each category for peak performance, we propose to directly learn the relationships among detections of different categories. As room geometry is also an important cue for object detection, we integrate Manhattan layout hypotheses for~\emph{holistic scene understanding}~\cite{song2015sun, lin2013holistic}.

Classically, structured prediction of spatial relationships is often accomplished via undirected~\emph{Markov random fields} (MRFs)~\cite{nowozin2011structured}.  As shown in Fig.~\ref{fig:exp_cascade}, this generally leads to a~\emph{fully connected} graph~\cite{rabinovich2007objects} because there are relationships among every pair of object categories.  An extremely challenging MAP estimation (or energy minimization) problem must then be solved at every training iteration, as well as for each test image, so learning and prediction are costly.  

We propose to instead adapt~\emph{cascaded classification}~\cite{heitz2009cascaded} to the modeling of contextual relationships in 3D scenes. In this approach, ``first-stage'' detections as in Sec.~\ref{sec:ssvm} become input features to ``second-stage'' classifiers that estimate confidence in the correctness of cuboid hypotheses.
This can be interpreted as a~\emph{directed} graphical model with hidden variables. Marginalizing the first-stage variables recovers a standard, fully-connected undirected graph. Crucially however, the cascaded representation is far more efficient: training~\emph{decomposes} into independent learning problems for each node (object category), and optimal test classification is possible via a rapid~\emph{sequence} of local decisions.

\subsection{Contextual Features}
For an overlapping pair of detected bounding boxes $B_i$ and $B_j$, we denote their volumes as $V(B_i)$ and $V(B_j)$, the volume of their overlap as $O(B_i, B_j)$, and the volume of their union as $U(B_i, B_j)$.  We characterize their geometric relationship via three features:
$S_1(i, j) = \frac{O(B_i, B_j)}{V(B_i)}$,
$S_2(i, j) = \frac{O(B_i, B_j)}{V(B_j)}$, and the intersection-over-union
$S_3(i, j) = \frac{O(B_i, B_j)}{U(B_i, B_j)}$.
To model contextual relations between objects and the scene layout $M$~\cite{lin2013holistic}, we compute the distance $D(B_i,M)$ and angle $A(B_i,M)$ of cuboid $B_i$ to the closest wall.

First-stage detectors provide a most-probable layout hypothesis, as well as a set of detections (following non-maximum suppression) for each category.
For a bounding box $B_i$ with confidence score $z_i$, there may be several overlapping bounding boxes of categories $c\in\{1, \ldots, C\}$.  Letting $i_c$ be the instance of category $c$ with maximum confidence $z_{i_c}$, features $\psi_i$ for bounding box $B_i$ are created via a quadratic function of $z_i$, $S_{1:3}(i, i_c)$, $A(B_i, M)$, and a radial basis expansion of $D(B_i, M)$.  Relationships between second-stage layout candidates and object cuboids are modeled similarly.

\begin{figure}[b]
\centering
\includegraphics[width=0.8\linewidth]{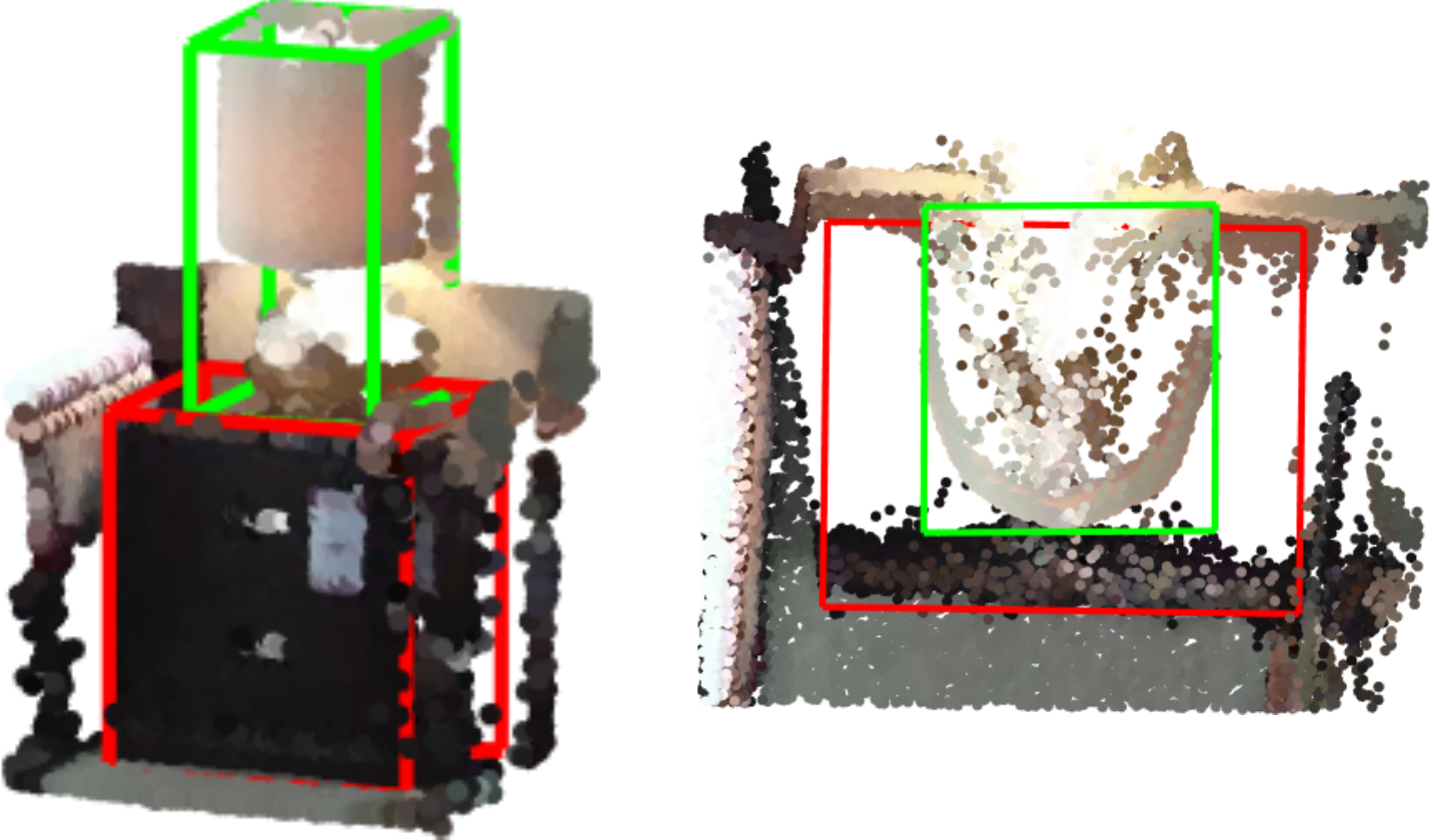}
  \caption{To model contextual relationships between small objects and the large objects supporting them, we compute the 2D areas and overlaps between 3D bounding boxes (left) seen from a top-down view (right).}
   \label{fig:2doverlap}
\end{figure}

For small objects that are placed on the support surfaces of large objects, 3D overlap features
are noisy. We replace 3D overlap with 2D overlap scores from the top-down view of the scene (Fig.~\ref{fig:2doverlap}). 
See the Appendix for further details.

\subsection{Contextual Learning}
Due to the directed graphical structure of the cascade, each second-stage detector may be learned independently.  The objective is a simple binary classification: is the candidate detection a true positive, or a false positive?
During training, each detected bounding box for each class is marked as ``true'' if its intersection-over-union score to a ground truth instance is greater than 0.25, and is the largest among such detections. We train a standard binary SVM with a radial basis function (RBF) kernel 
\begin{equation}
  K(B_i,B_j) = \exp \left(-\gamma ||\psi_i-\psi_j||^2\right).
\end{equation}
The bandwidth parameter $\gamma$ is chosen using validation data.  While we use a RBF kernel for all reported experiments, the performance of a linear SVM is only slightly worse, and cascaded classification still provides useful performance gains for that more scalable training objective. 

To train the second-stage layout predictor (the bottom node in Fig.~\ref{fig:exp_cascade}), we combine the object-layout features with the Manhattan voxel features from Sec.~\ref{sec:layout}, 
and again use S-SVM training to optimize the free-space IOU.

\subsection{Contextual Prediction}
During testing, given the set of cuboids found in the first-stage sliding-window search, we apply the second-stage cascaded classifier to each cuboid $B_i$ to get a new contextual confidence score $z'_i$.  The overall confidence score used for precision-recall evaluation is then $z_i + z'_i$, to account for both the original belief from the geometric and COG features and the correcting power of contextual cues. The second-stage layout prediction is directly provided by the second-stage S-SVM classifier.

\section{Experiments}
\label{sec:experiments}
We train our 3D object detection algorithm solely on the SUN RGB-D dataset~\cite{song2015sun} with 5285 training 
images, and report performance on 5050 test images for all 19 object categories (Table~\ref{table:ap_score}). 
The NYU Depth dataset~\cite{silberman2012indoor} has 3D cuboid labels for 1449 images, but annotations are noisy and inconsistent.
Some previous work has only evaluated detection performance on this small dataset~\cite{guptaCVPR15align}, or defined their 
own annotations for 3D cuboids~\cite{zhuo17amodal3d}. We do not evaluate on the NYU Depth dataset because it is a subset of SUN RGB-D.

We evaluate detection performance via the \emph{intersection-over-union} (IOU) with ground-truth cuboid annotations, and consider the predicted cuboid to be correct when the IOU is above 0.25. To evaluate the layout prediction performance, we calculate the free space IOU with human annotations. We provide results demonstrating the effectiveness of our 3D scene understanding system, and the importance of both appearance and context features.

\subsection{Modeling Latent Support Surfaces}
\begin{figure*}[!ht]
  \centering
\begin{tabular}{cc|cc}
 \multicolumn{4}{c}{\includegraphics[width=0.95\textwidth]{figs/exp_detections/legend.pdf}} \\
\includegraphics[width=.225\textwidth]{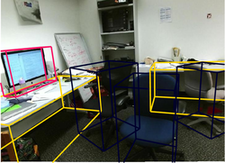}&
\includegraphics[width=.225\textwidth]{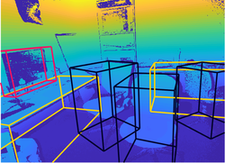}&
\includegraphics[width=.225\textwidth]{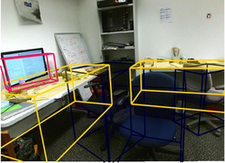}&
\includegraphics[width=.15\textwidth]{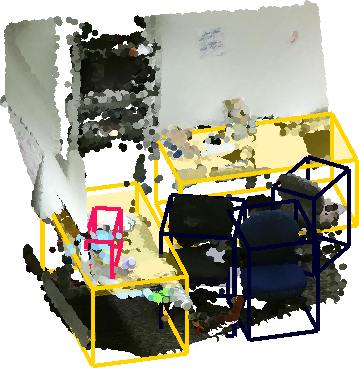}\\
\includegraphics[width=.225\textwidth]{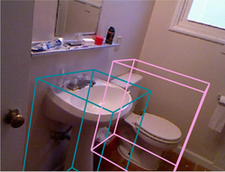}&
\includegraphics[width=.225\textwidth]{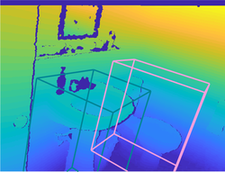}&
\includegraphics[width=.225\textwidth]{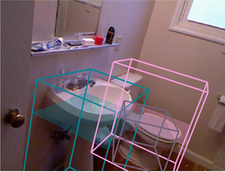}&
\includegraphics[width=.15\textwidth]{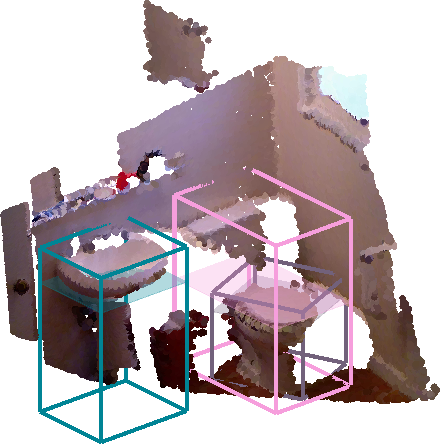}\\
\includegraphics[width=.225\textwidth]{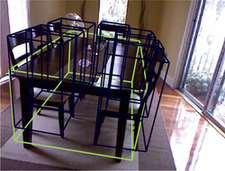}&
\includegraphics[width=.225\textwidth]{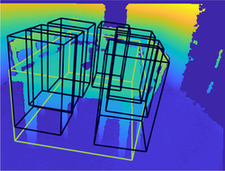}&
\includegraphics[width=.225\textwidth]{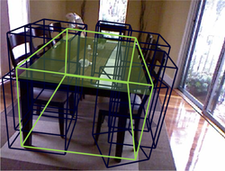}&
\includegraphics[width=.225\textwidth]{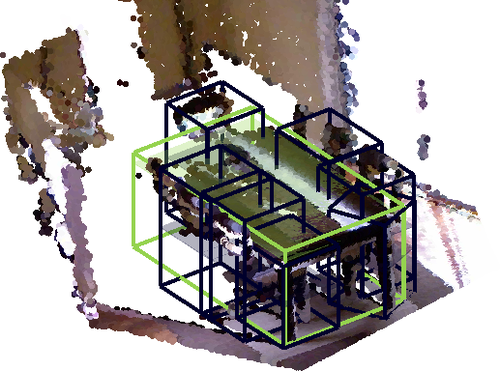}\\
\includegraphics[width=.225\textwidth]{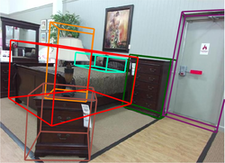}&
\includegraphics[width=.225\textwidth]{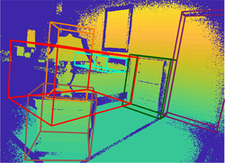}&
\includegraphics[width=.225\textwidth]{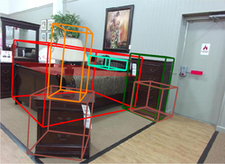}&
\includegraphics[width=.200\textwidth]{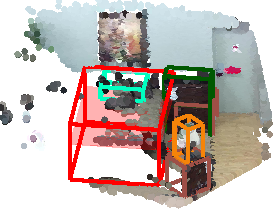}\\
\includegraphics[width=.225\textwidth]{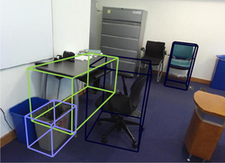}&
\includegraphics[width=.225\textwidth]{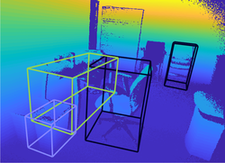}&
\includegraphics[width=.225\textwidth]{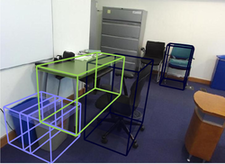}&
\includegraphics[width=.225\textwidth]{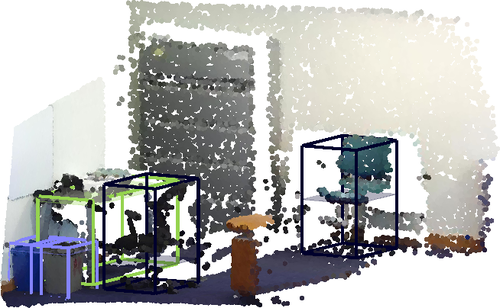}\\
\includegraphics[width=.225\textwidth]{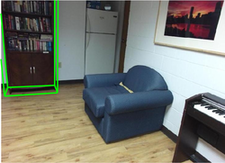}&
\includegraphics[width=.225\textwidth]{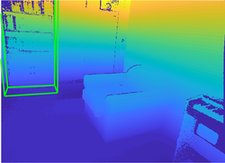}&
\includegraphics[width=.225\textwidth]{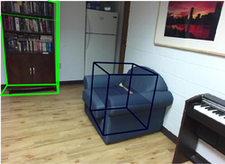}&
\includegraphics[width=.225\textwidth]{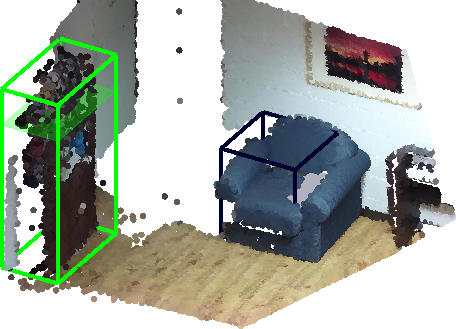}\\
\includegraphics[width=.225\textwidth]{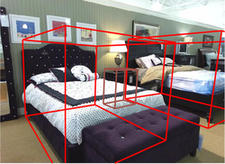}&
\includegraphics[width=.225\textwidth]{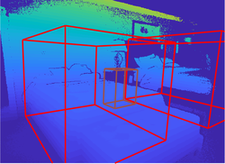}&
\includegraphics[width=.225\textwidth]{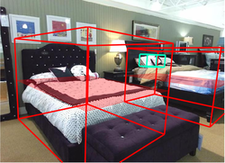}&
\includegraphics[width=.225\textwidth]{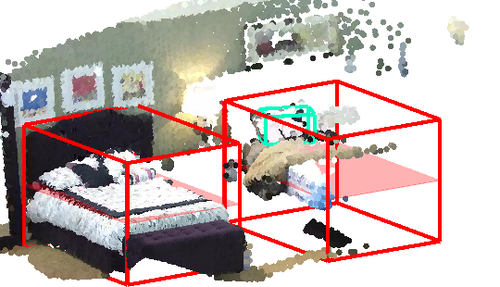}\\
\multicolumn{2}{c}{Groundtruth Annotations for RGB-D Images} &\multicolumn{2}{c}{Our Final Stage 3D Detection Output}\\
  \end{tabular}
  \caption{Visualizing our final stage 3D detections for objects with high confidence scores. 
	  Support surfaces are depicted with faded colors inside each large object.}
    \label{fig:vis_3ddetection_results}
\end{figure*}

For objects such as beds, tables, and desks, modeling support surface as a latent variable helps capture the intra-class
style variations within each cuboid. 
We visualize examples of inferred support surfaces in Figure~\ref{fig:vis_3ddetection_results}.
For objects that do not have explicit ``support surfaces'', such as bathtub, bookshelf, and sink, our model can be 
viewed as a single part-based model and is also effective for 3D object detection.  Note that the goal of this work 
is to model latent support surfaces to boost 3D detection accuracy, not to predict accurate
supporting regions in scenes. 
We do not use any annotations of support surfaces when training, and also do not evaluate our performance on surface prediction benchmarks~\cite{guo2013support}.

\begin{figure}[!t]
  \centering
  \includegraphics[width=.45\textwidth]{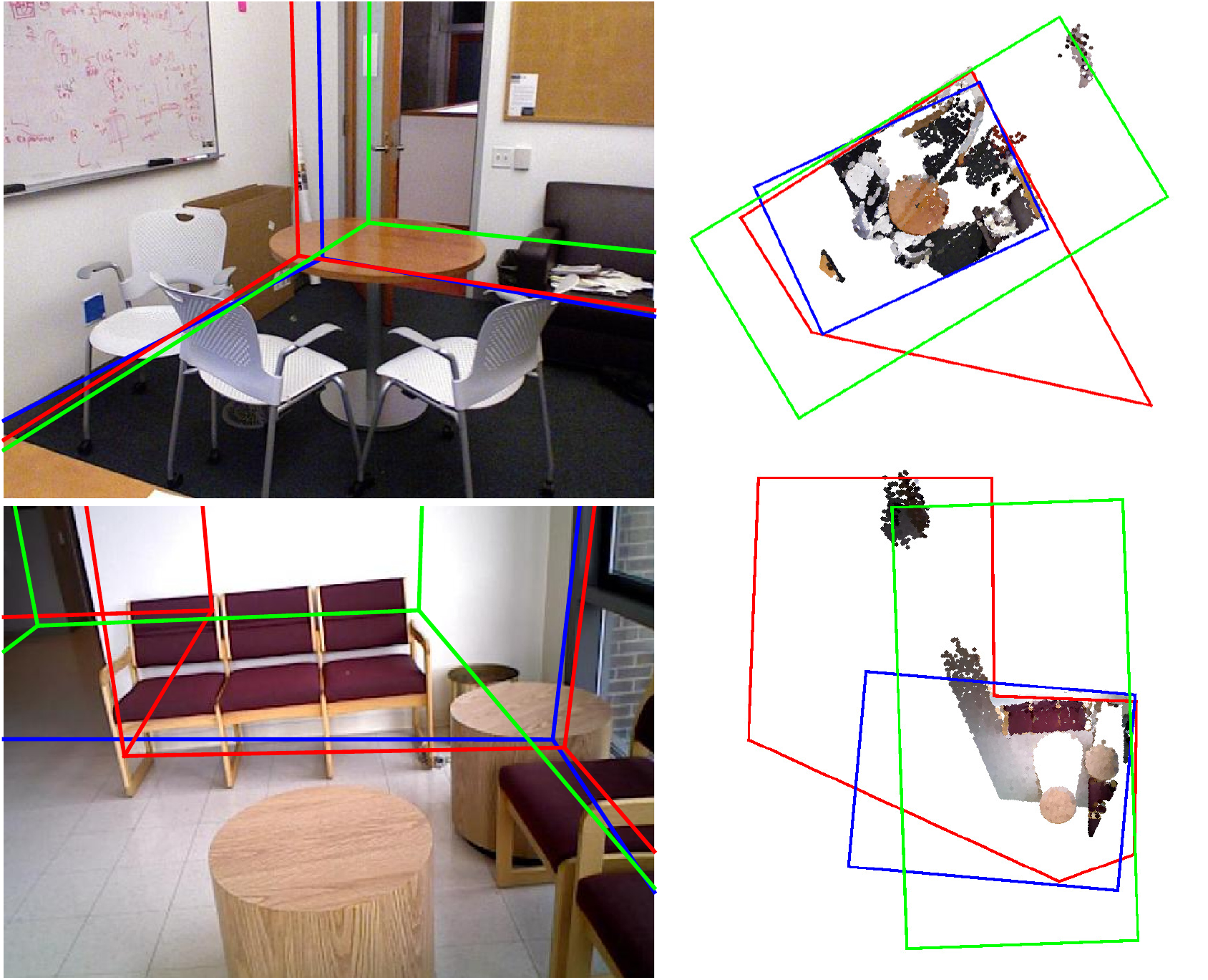}
\caption{Comparison of our Manhattan voxel 3D layout predictions ({blue}) to the SUN RGB-D baseline~\cite{song2015sun} (green) and the ground truth annotations (red).  Our learning-based approach is less sensitive to outliers and degrades gracefully in cases where the true scene structure violates the Manhattan world assumption.}
  \label{fig:vis_layout}
\end{figure}

\subsection{Small Object Detection}
Detecting small objects is a challenging task, and achieving high accuracy remains an open research problem. 
Without modeling support surfaces, our baseline detectors completely fail to detect small objects because the search space is large and
3D object proposals contain many false positives. Using simple heuristics to check support relationships in the 
SUN-RGBD annotations, we find that more than 95\% of lamps/pillows/monitors/TVs are placed on the surface of night-stands/tables/beds/desks/dressers. 
As shown in Table~\ref{table:ap_score}, searching on predicted surfaces thus enables our algorithm to discover small objects with higher precision.

\subsection{The Importance of Context}
\begin{figure*}[!ht]
  \centering
  \vspace{-1em}
  \begin{tabular}{ccccc}
\includegraphics[width=.20\textwidth]{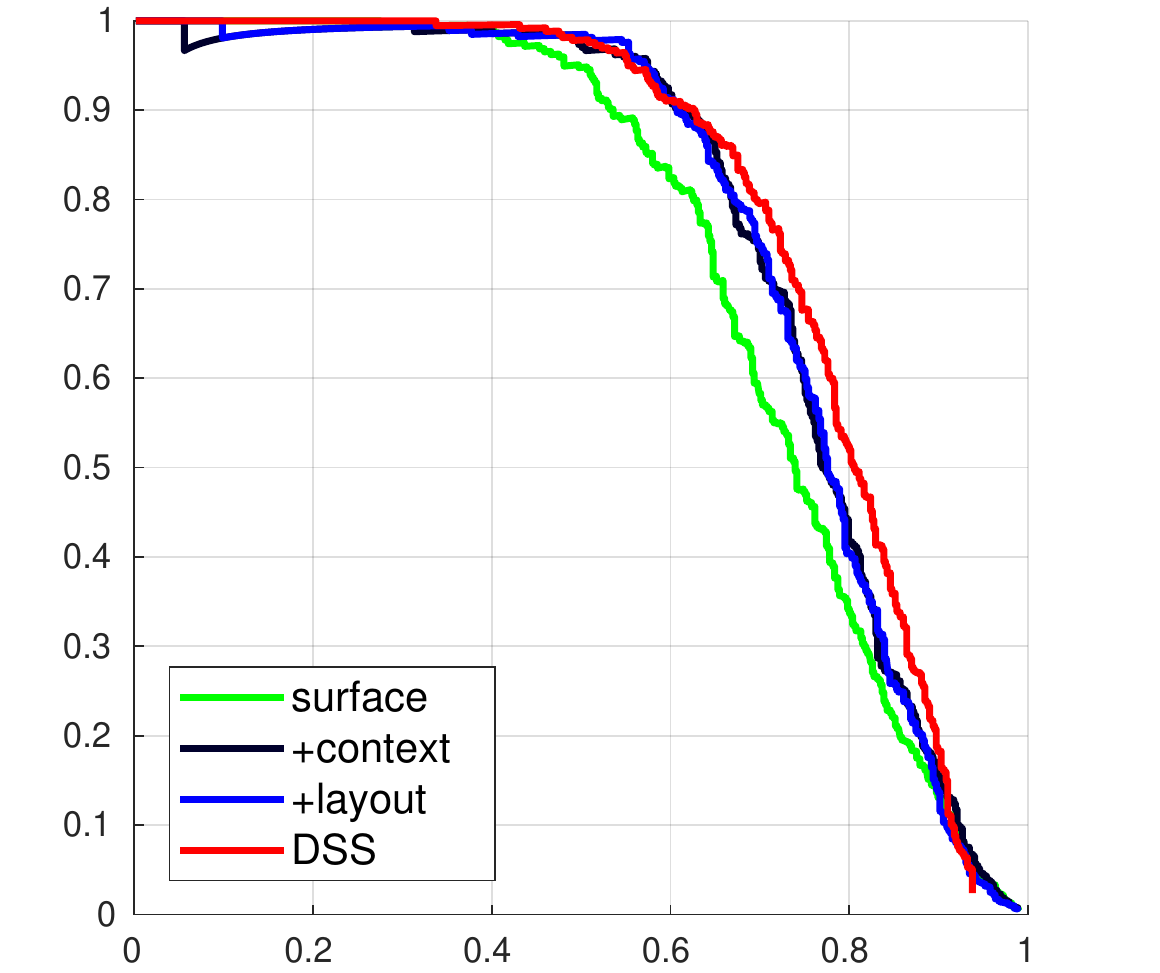}&
    \includegraphics[width=.20\textwidth]{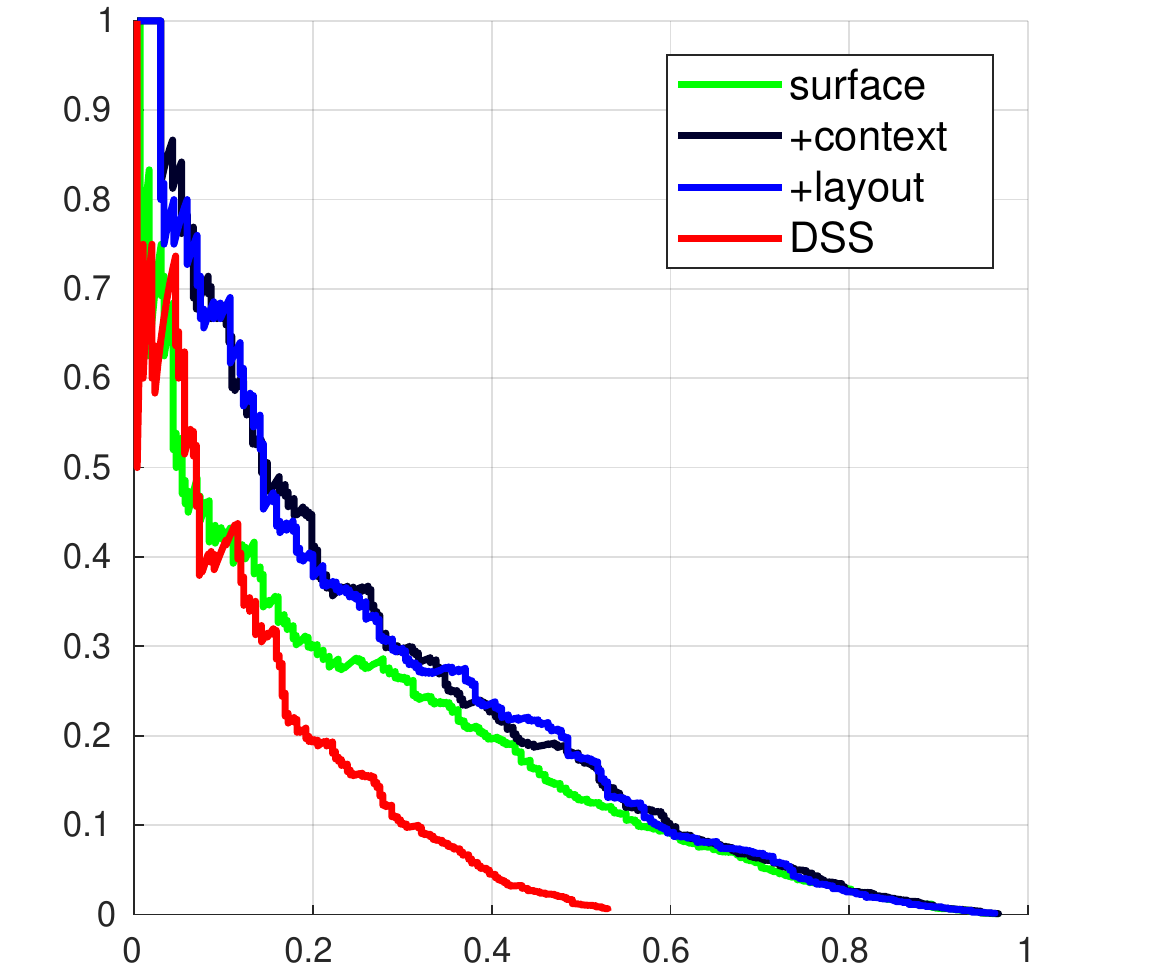}&
    \includegraphics[width=.20\textwidth]{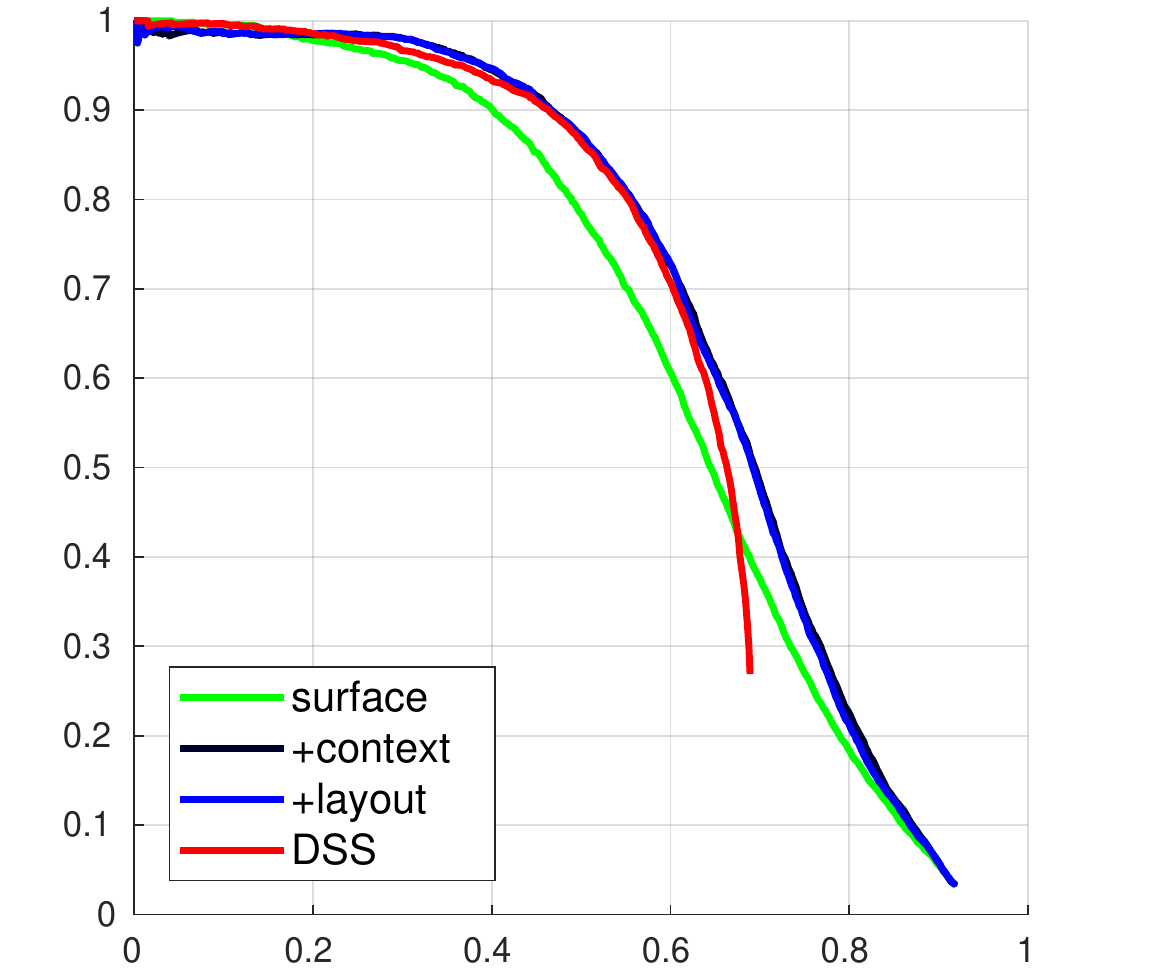}&
    \includegraphics[width=.20\textwidth]{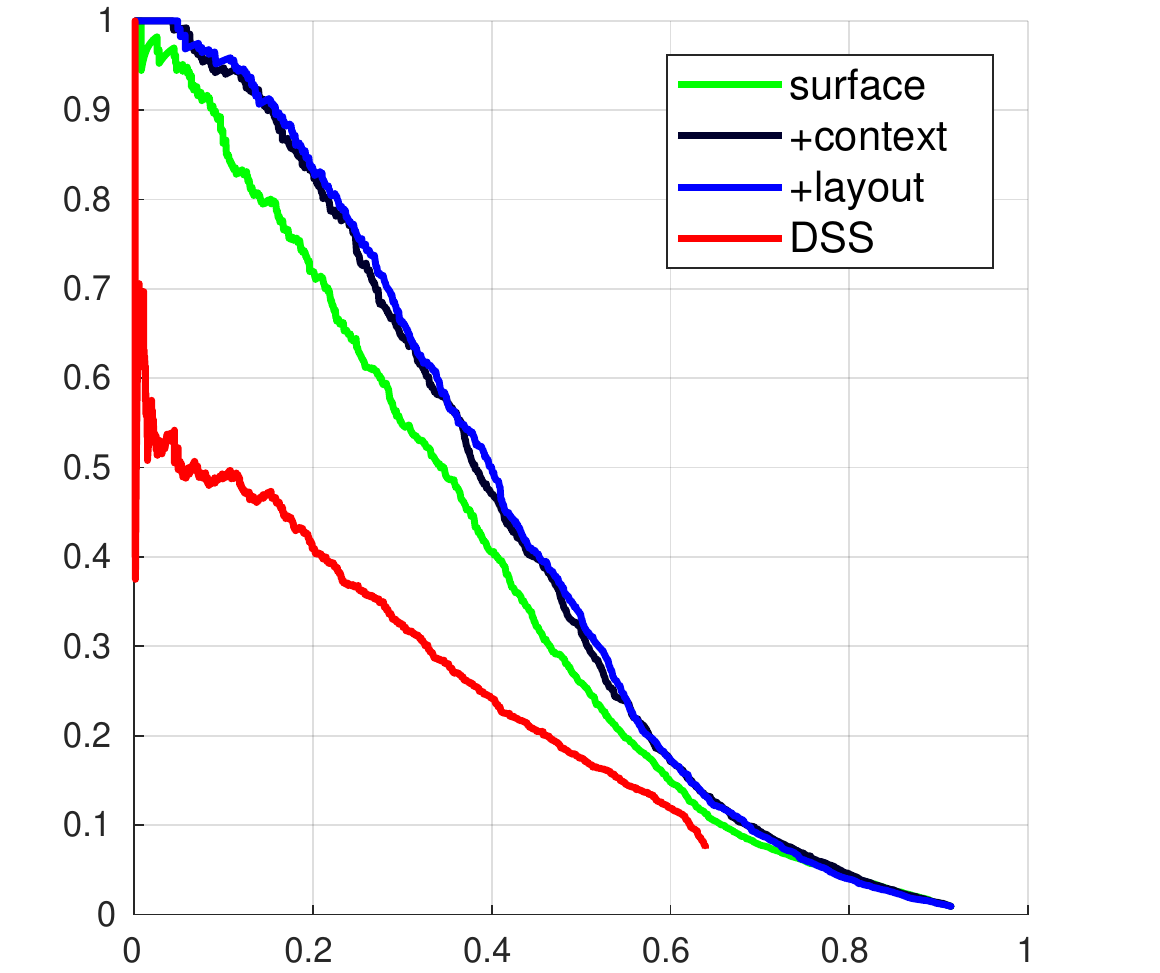}&
    \includegraphics[width=.20\textwidth]{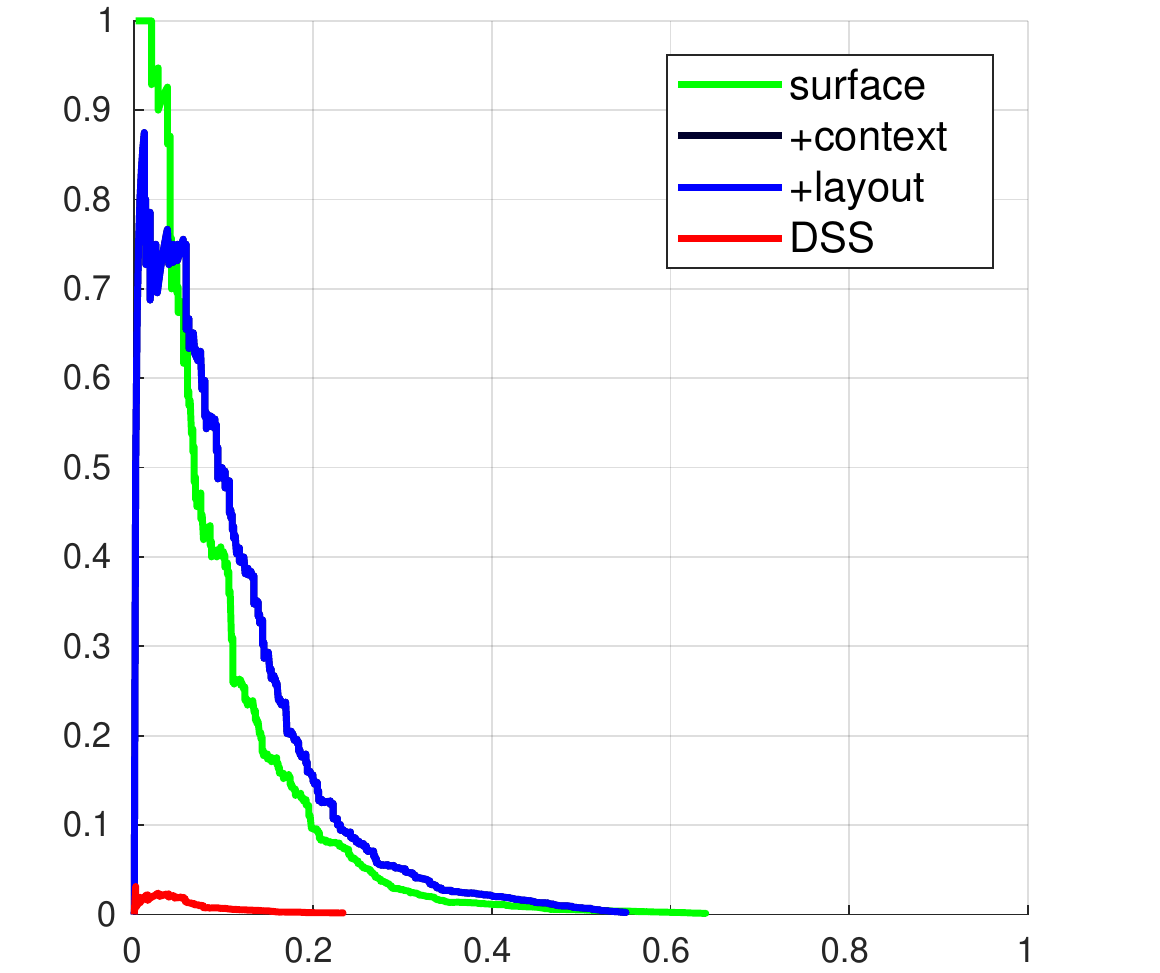} \\
 \small \hspace{-0em}  Bed& \small \hspace{-0em}Bookshelf& \small  \small \hspace{-0em}Chair& \small \hspace{-0em}Desk& \small \hspace{-0em}Monitor \\
  \end{tabular}
  \caption{ Precision-recall curves for several object categories, including monitors which are supported by the surfaces of other objects, on the SUN RGB-D dataset~\cite{song2015sun}.  We compare our \cog detector with latent support surfaces, and possibly also context and layout cues, to the \emph{deep sliding shape} (DSS) method~\cite{song2016deep}.}
  \label{fig:pr_curve}
\end{figure*}

To show that the cascaded classifier helps to prune false positives, we evaluate detections using 
the confidence scores from the first-stage classifier (surface), as well as the updated confidence scores 
from the second-stage classifier using all object-to-object features (+context). 
As shown in Table~\ref{table:ap_score} and Fig.~\ref{fig:pr_curve}, 
adding a contextual cascade clearly boosts performance. 
Furthermore, when object-to-scene-layout features are included (+layout), 
performance increases further. This result demonstrates that even if a small 
number of object categories are of primary interest, building models of the broader scene can be very beneficial.

\begin{figure*}[!ht]
  \centering
\begin{tabular}{cccc}
 \multicolumn{4}{c}{\includegraphics[width=0.95\textwidth]{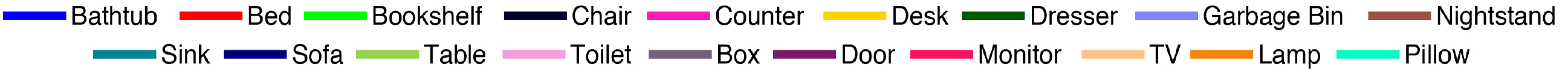}} \\
\includegraphics[width=.225\textwidth]{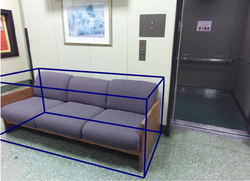}&
\includegraphics[width=.225\textwidth]{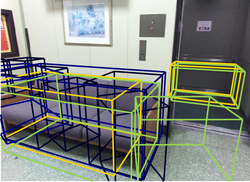}&
\includegraphics[width=.225\textwidth]{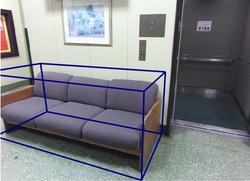} &
\includegraphics[width=.225\textwidth]{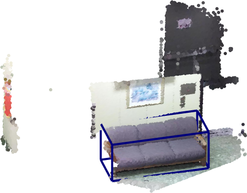}\\
\includegraphics[width=.225\textwidth]{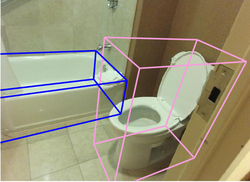}&
\includegraphics[width=.225\textwidth]{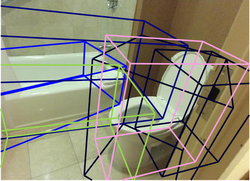}&
\includegraphics[width=.225\textwidth]{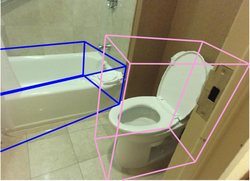} &
\includegraphics[width=.225\textwidth]{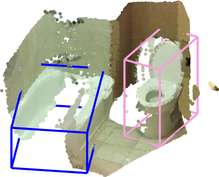}\\
{\small Ground Truth} &\hspace{-0em}{\small First-stage}&\hspace{-0em} {\small Second Stage} &\hspace{-0em} {\small Second Stage}
  \end{tabular}
  \caption{Detections with confidence scores larger than the same threshold for each stage of our cascaded classification framework. Notice that using contextual information helps prune away false positives while preserving true positives.}
    \label{fig:tp_example}
\end{figure*}

We show some representative detection results in Fig.~\ref{fig:tp_example}. 
In the first image our chair detector is confused and fires on part of the sofa, 
but with the help of contextual cues of other detected bounding boxes, these
false positives are pruned away. 
For a fixed threshold across all object categories, we have as many 
true detections while producing fewer false positives.

\subsection{Cubical Voxels versus Manhattan Voxels}
We use the free-space IOU~\cite{song2015sun} to evaluate layout prediction performance.  Using standard cubical voxels, our performance (\textbf{72.33}) is similar to the heuristic SUN RGB-D baseline (\textbf{73.4},~\cite{song2015sun}).  Combining Manhattan voxels with structured learning, performance increases to \textbf{78.96}, demonstrating the effectiveness of  this improved discretization. Furthermore, if we also incorporate contextual cues from detected objects, the score improves to \textbf{80.03}. We provide layout prediction examples in Fig.~\ref{fig:vis_layout}.

\newcommand{\specialcell}[2][c]{%
  \begin{tabular}[#1]{@{}c@{}}#2\end{tabular}}
  
\setlength{\tabcolsep}{0.05em}
\begin{table*}[ht!]
\centering
\begin{tabular}{c|ccccccccccccccccccc|cc} 
&Bathtub&Bed&Bookshelf&Chair&Desk&Dresser&Nightstand&Sofa&Table&Toilet&Box&Door&Counter&Garbage-bin&Sink&Pillow&Monitor&TV&Lamp&\specialcell[c]{mAP\\(10)}&\specialcell[c]{mAP\\(19)} \\ \hline
surface    &69.9&73.2&19.0&63.2&35.9&19.9&26.0&58.1&46.7&81.9&12.6&3.4&5.7&30.4&30.8&8.3&10.0&1.6&23.7&49.4&32.6\\ 
+context   &71.3&76.8&24.8&\textbf{67.4}&40.4&22.8&39.0&60.5&51.4&85.6&14.4&4.1&12.6&34.7&38.7&7.7&11.4&2.1&24.0&54.0&36.3\\ 
+layout    &72.0&76.8&25.5&67.2&41.0&23.7&39.7&60.4&51.7&85.6&14.4&4.1&\textbf{12.7}&\textbf{34.9}&39.4&7.6&\textbf{11.4}&2.1&\textbf{25.4}&\textbf{54.3}&\textbf{36.6}\\ \hline
DSS~\cite{song2016deep}&44.2&{78.8}&11.9&61.2&20.5&6.4&15.4&53.5&50.3&78.9&1.5&0.0&4.1&20.4&32.3&{13.3}&0.2&0.5&{18.4}&42.1&26.9\\
C3D~\cite{yuzhuo2018C3D}&60.3&\textbf{82.9}&\textbf{33.9}&63.7&22.1&18.5&30.6&56.5&\textbf{57.3}&85.7&18.5&5.0&10.1&25.7&35.0&16.1&4.7&\textbf{4.8}&25.3&51.2&34.6\\
Ren2018~\cite{ren2018latent}&\textbf{76.2}&73.2&{32.9}&60.5&34.5&13.5&30.4&{60.4}&{55.4}&73.7&\textbf{19.5}&\textbf{5.4}&{10.7}&{34.6}&\textbf{75.3}&\textbf{12.5}&{1.6}&{2.1}&16.9&{51.0}&{36.3}\\
Ren2016~\cite{ren2016cog}&58.3&63.7&31.8&{62.2}&\textbf{45.2}&15.5&27.4&51.0&51.3&70.1&-&-&-&-&-&-&-&-&-&47.6&-\\
Lahoud~
\cite{lahoud20172d}&43.5&64.5&31.4&48.3&27.9&{25.9}&{41.9}&40.4&37.0&{80.4}&-&-&-&-&-&-&-&-&-&45.1&-\\
Frustum~\cite{qi2017frustum}&43.3&81.1&33.3&64.2&24.7&\textbf{32.0}&\textbf{58.1}&\textbf{61.1}&51.1&\textbf{90.9}&-&-&-&-&-&-&-&-&-&54.0&-\\
SS~\cite{song2014sliding}&-&43.0&-&28.2&-&-&-&20.6&19.7&60.9&-&-&-&-&-&-&-&-&-&-&-\\
\end{tabular}
\caption{Experimental results on the SUN RGB-D dataset~\cite{song2015sun}. Modeling support surfaces (+surface) simultaneously helps detect
large objects and reduces false positives for small objects (last 4 categories). 
The final stage of the cascaded classifier (+cascade)
models object context and possibly also layout context (+layout). These cues reduce false positives and boost 
average performance to the state-of-the-art for the first 10, as well as all 19, object categories.}
\label{table:ap_score}
\end{table*}

\begin{figure*}[!ht]
  \centering
  \begin{tabular}{ccccc}
    \hspace{0em}\includegraphics[width=.2\textwidth]{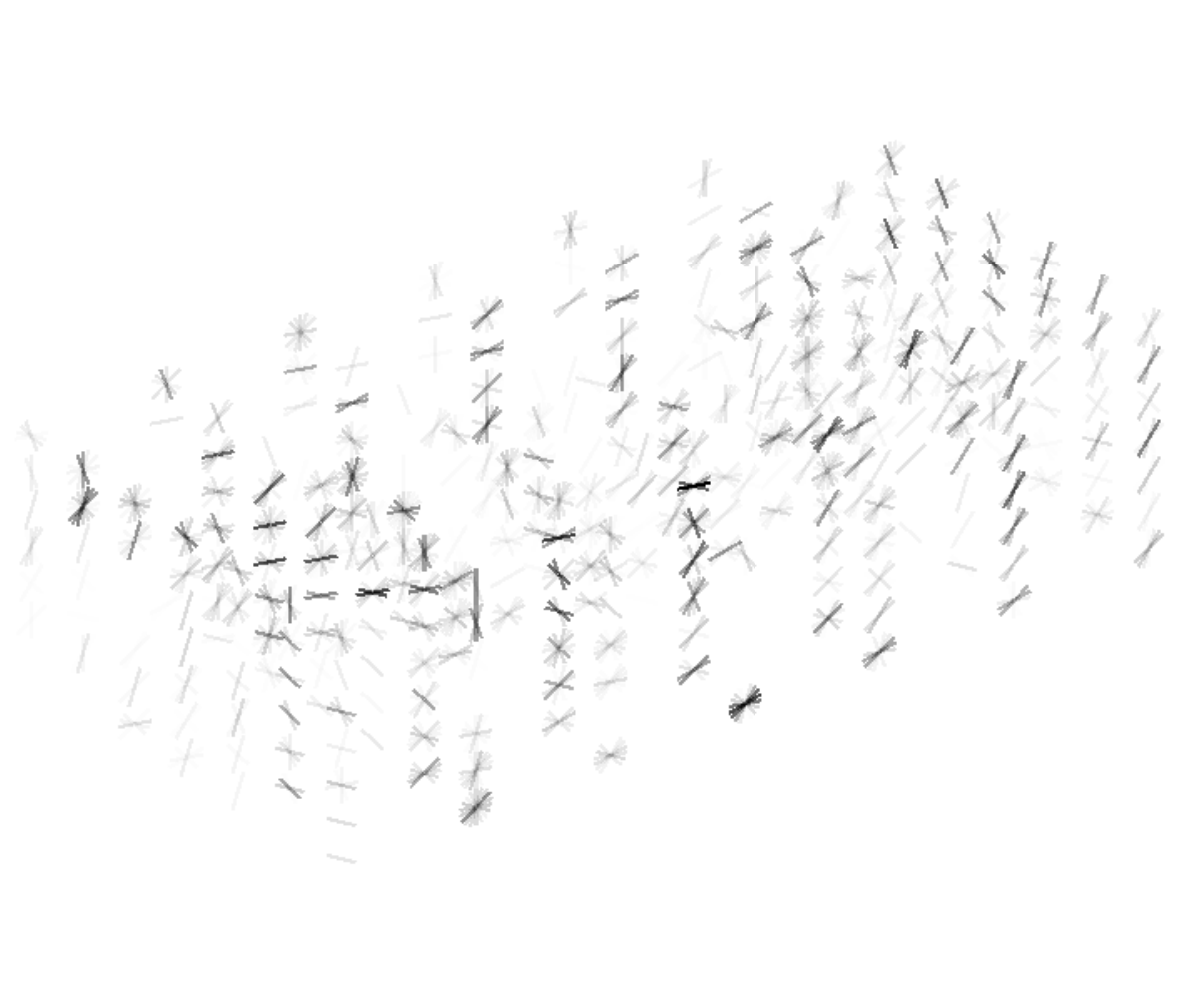}&
    \hspace{-1.5em}\includegraphics[width=.2\textwidth]{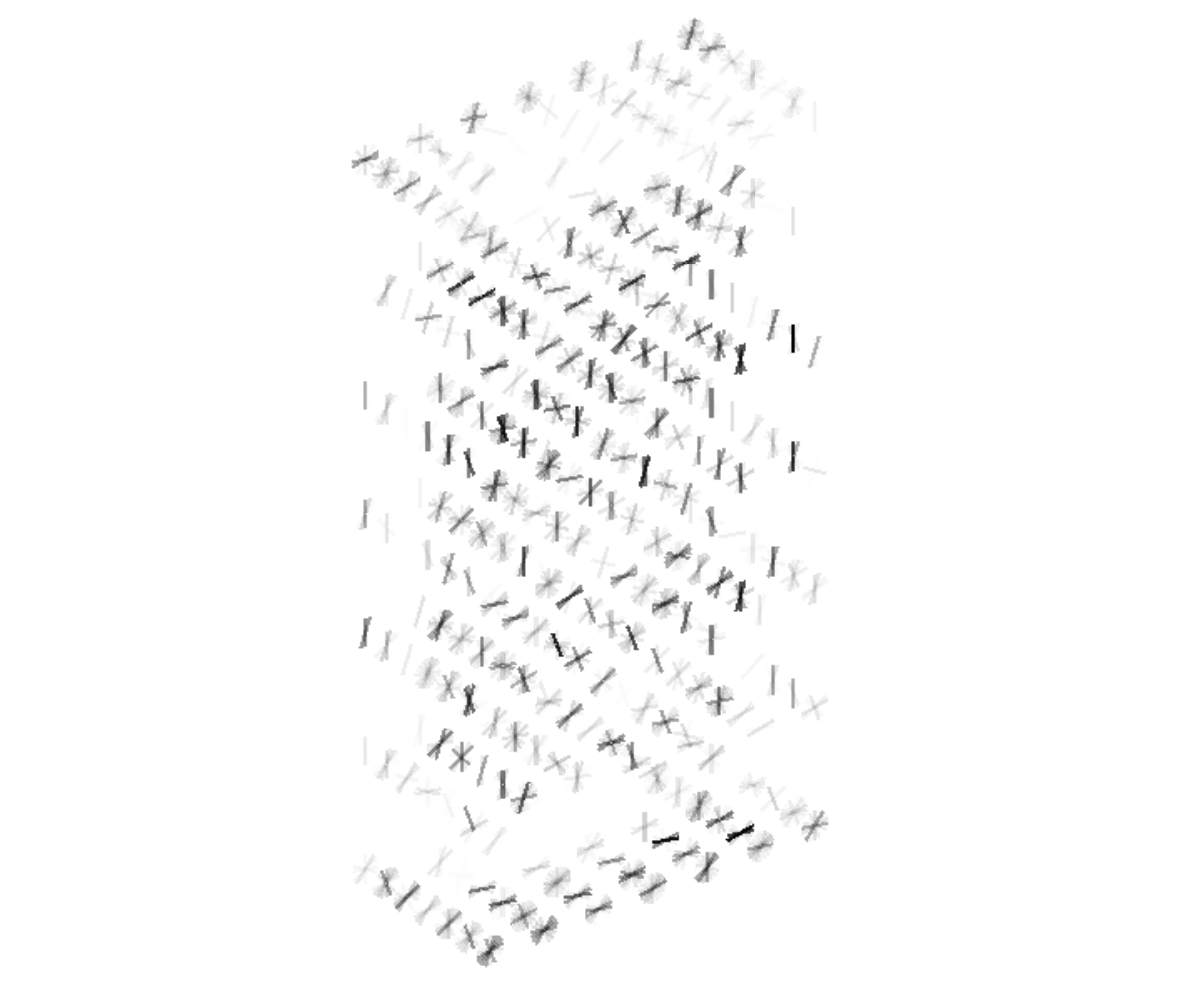}&
    \hspace{-1.5em}\includegraphics[width=.2\textwidth]{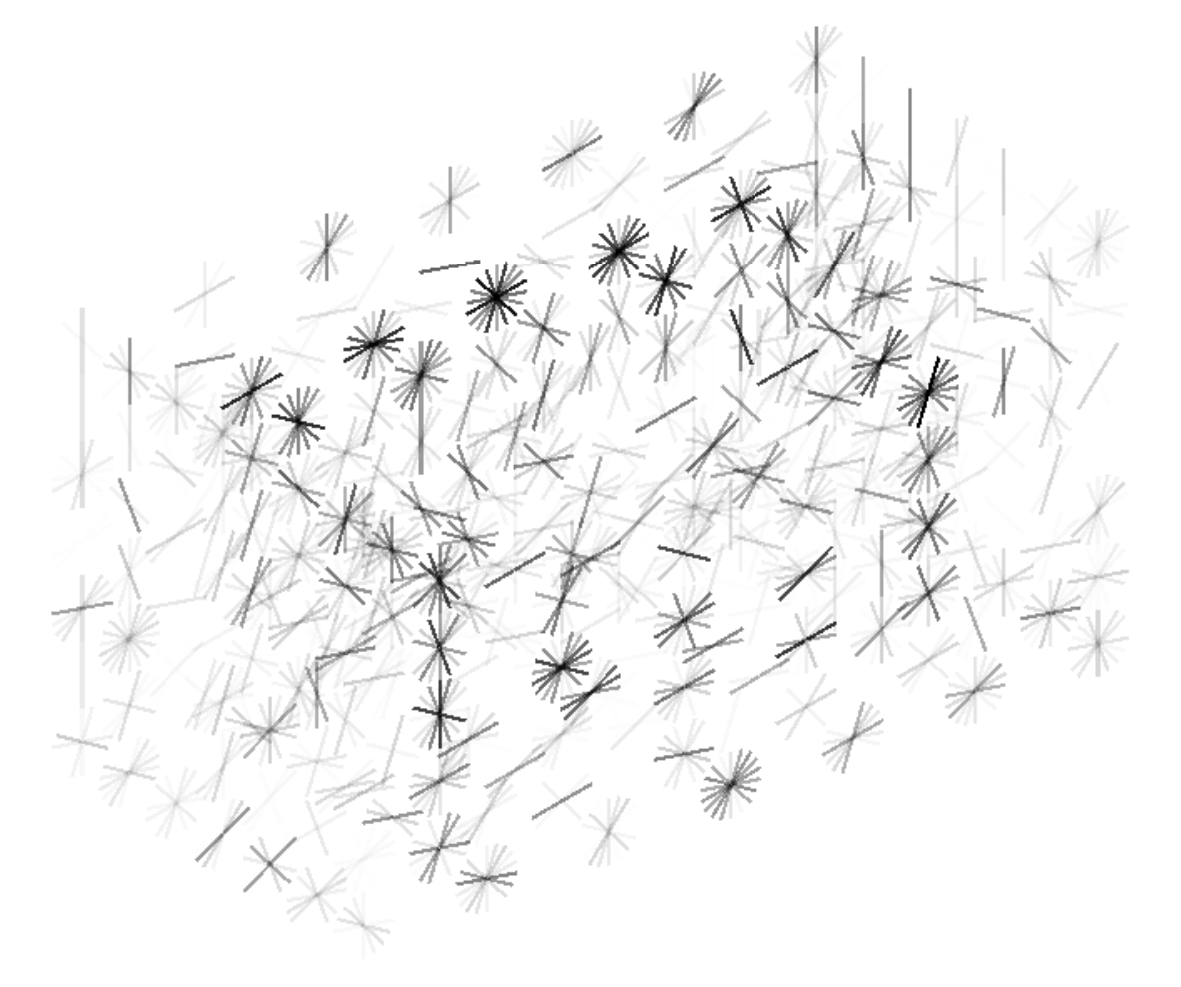}&
    \hspace{-1.5em}\includegraphics[width=.2\textwidth]{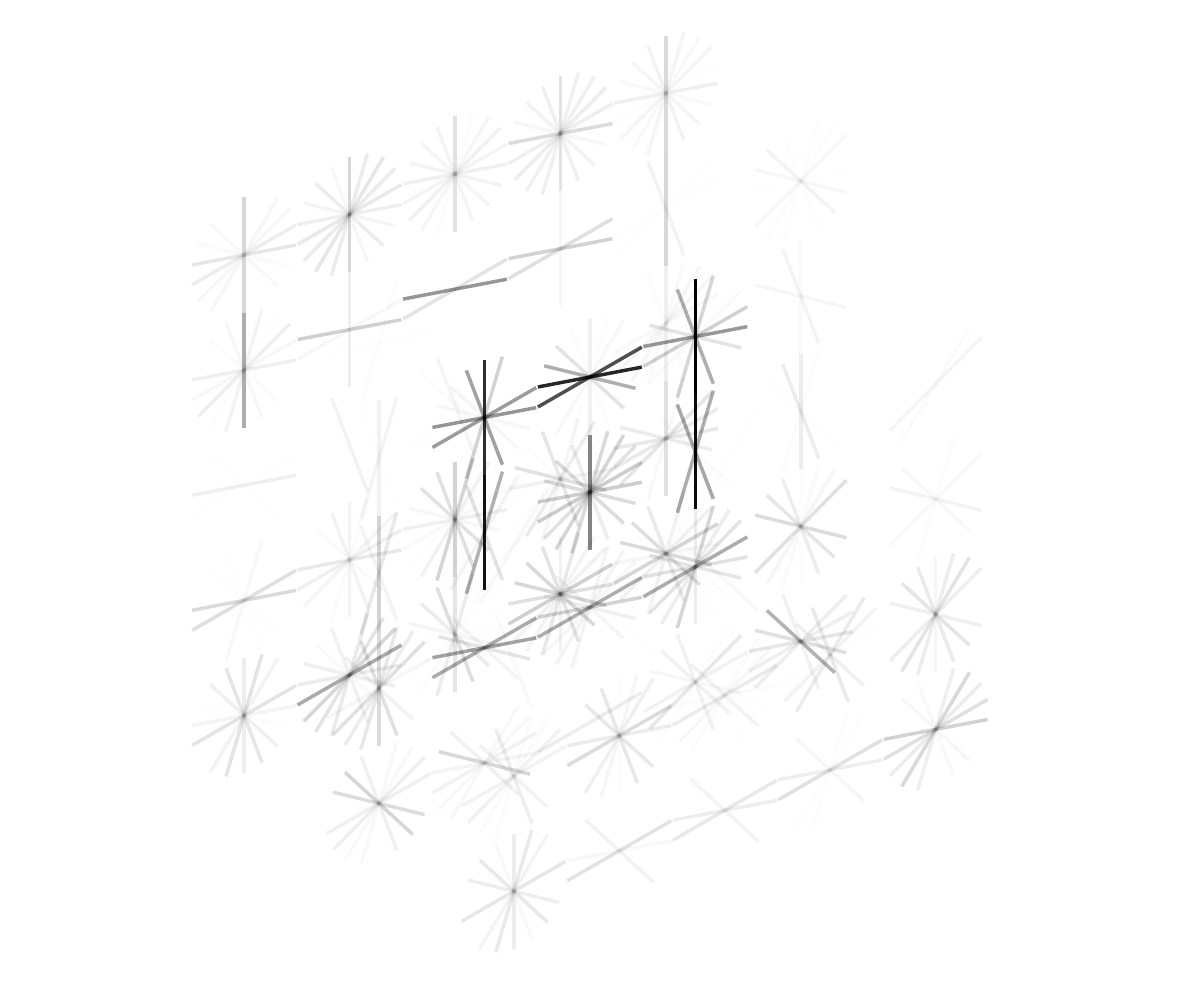}&
    \hspace{-1.5em}\includegraphics[width=.2\textwidth]{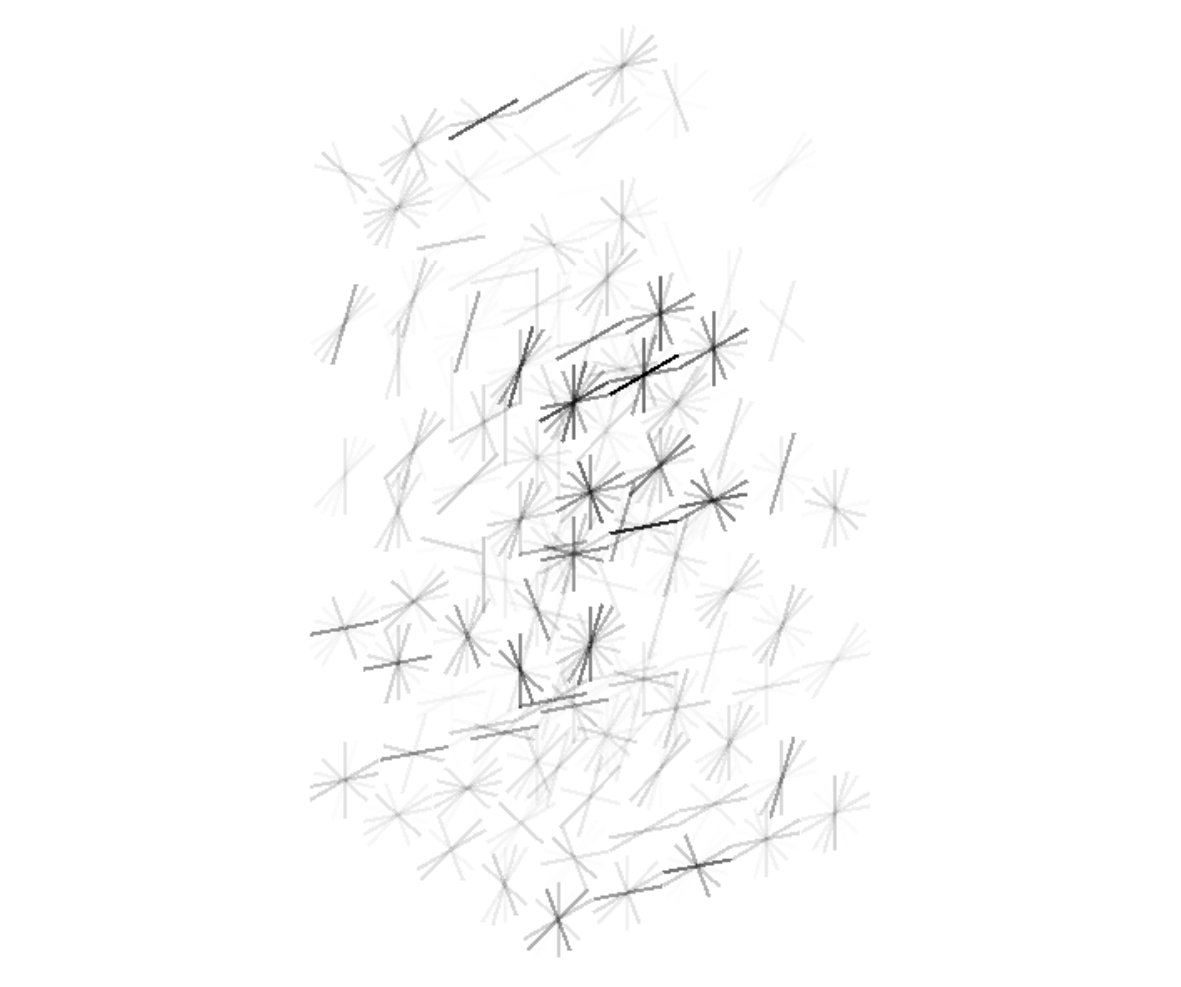} \\
    \small Bathtub&\small\hspace{-1.5em}Bookshelf&\small\hspace{-1.5em}Sofa&\small\hspace{-1.5em}Monitor&\small\hspace{-1.5em}Lamp\\
  \end{tabular}
  \caption{Visualization of learned 3D \cog descriptors with expanded cuboid features for several categories. 
  Reference orientation bins with larger weights are darker, providing a 3D visualization of the typical appearance of each object category. Cuboid sizes are matched to the median of the training data.}
  \label{fig:vis_cog}
\end{figure*}

\subsection{Computational Speed}
We implemented our algorithm using MATLAB in a 2.5GHz single core CPU. The computational speed of 
our detector is 10-30min per image. The most time-consuming
part is the feature computation step, which could be improved by using parallel computing with multi-core CPUs or GPUs.
With pre-computed cuboid features for each RGB-D image, the inference time is 2sec for each object category. 
With pre-computed contextual features among all objects, the cascaded prediction framework takes less than 0.5sec on average.
The training time ranges from 2 to 12 hours per category, depending on the number of training instances. 

Other deep learning-based 3D detection systems~\cite{song2016deep,qi2017frustum} typically 
have a region proposal step that highly constrains the search space for all object categories. 
Our cuboid proposals are dense and extensive, and thus the computational speed is usually 
slower. This limitation of our system could be potentially alleviated by pre-processing the data
using a region proposal network~\cite{song2016deep}.

\subsection{Comparison to Other Methods}
This paper has several differences from our preliminary work~\cite{ren2016cog,ren2018latent}. 
Our use of expanded cuboid features is new, and contributes to our overall 3D detection performance.
Some implementation details also differ, for example~\cite{ren2018latent} uses scene category features while this paper does not.  Also~\cite{ren2016cog} 
uses a $6\times6\times6$ discretization of cuboids into voxels, and uses only images containing at least one object instance for structural SVM training of detectors.

Compared to other methods that use CNN features~\cite{song2016deep,lahoud20172d} pretrained
on external datasets, our COG-based 3D object detector has comparable or better performance even without the contextual cues provided by our cascaded classifier.
Conventional CNNs for 3D detection~\cite{song2016deep,lahoud20172d} are trained to produce weighted confidence 
scores for each of multiple object categories, while our first-stage detector is instead tuned to discriminatively localize individual categories in 3D.  
Our subsequent cascaded prediction~\cite{heitz2009cascaded} of contextual relationships between object detections
has structural similarities to a multi-stage neural network, but it is trained using (convex) structural SVM loss functions and designed to have a more interpretable, graphical structure.
Interestingly, our overall cascaded approach is more accurate than standard 3D CNNs~\cite{song2016deep,lahoud20172d,qi2017frustum} in the detection of both 10 and 19 object categories.

\setlength\tabcolsep{3pt} 

\section{Conclusion}

We propose a geometric framework for 3D cuboid detection and Manhattan layout prediction from RGB-D images.  Using our novel \cog descriptor of 3D cuboid appearance, we train accurate 3D object detectors for nineteen categories, as well as a cascaded classifier that learns contextual cues to boost performance. 
Modeling the height of support surfaces as latent variables further increases detection accuracy for large objects, and constrains the search space to make the detection of small objects feasible.

Our scene representations are learned directly from RGB-D data without external CAD models, and thus may be easily generalized to many other object categories.
Gradient-based detectors incorporating cloud of oriented gradient (COG) features achieve state-of-the-art performance on the challenging SUN RGB-D dataset.
We hypothesize that our improvement over baseline methods incorporating deep learning is due to the superior ability of \cog descriptors to generalize to novel 3D viewpoints.  Incorporating similar geometric invariances into convolutional networks is a promising area for future research.

\newpage
\appendix[Computation of Contextual Features]
\label{sec:cascade_feature}
We give a more detailed specification of the contextual features we use to model object-object and object-layout relationships.

The first-stage detectors provide a most-probable layout hypothesis, as well as a set of detections (following non-maximum suppression) for each category. For each bounding box $B_i$ with confidence score $z_i$, there may be several bounding boxes of various categories $c\in\{1, 2, ..., C\}$ that overlap with it. We let $i_c$ be the instance of category $c$ with the maximum confidence score $z_{i_c}$. The features $\psi_i$ for bounding box $B_i$ are then as follows:
\begin{enumerate}
\setlength{\itemsep}{0pt}
\item Constant bias feature, and confidence score $z_i$ from the first-stage detector.
\item For $m\in\{1,2,3\}$, $c\in\{1, 2, ..., C\}$, we calculate $S_m(i, i_c)$, $S_m(i, i_c)\cdot z_{i_c}$, $S_m(i, i_c)\cdot z_{i}$ and concatenate those numbers.
\item For $c\in\{1, 2, ..., C\}$, we calculate the difference in confidence score from each first-stage detector, $z_i - z_{i_c}$, and concatenate those numbers.
\item For $D(B_i, M)$, we consider radial basis functions of the form in Eq.~\ref{eq:rbf}.
For a typical indoor scene, the largest object-to-wall distance is usually less than $5$m, therefore we space the basis function centers $\mu_j$ evenly between 0 and 5 with step size 0.5, and choose $\sigma=0.5$. We expand $D(B_i, M)$ using this radial basis expansion.
\item The absolute value of cosine $D(B_i, M)$: $|\cos(D(B_i, M))|$.
\end{enumerate}

To model the second-stage layout candidates, we select the bounding box $i_c$ with the highest confidence score $z_{i_c}$ from the first-stage classifier in each category $c\in\{1, 2, ..., C\}$, and use the following features for layout $M_i$ with confidence score $z'_i$:
\begin{enumerate}
\setlength{\itemsep}{0pt}
\item All the features used in the first-stage to model $M_i$ using \emph{Manhattan Voxels}.
\item For $c\in\{1, 2, ..., C\}$, we calculate the radial basis expansion for $D(B_{i_c}, M_i)$, and its product with $z'_i$ and $z_{i_c}$.
\item For $c\in\{1, 2, ..., C\}$, we calculate the absolute value of the cosine of $D(B_{i_c}, M_i)$: $|\cos(D(B_{i_c}, M_i))|$,  $|\cos(D(B_{i_c}, M_i))|\cdot z'_i$ and $|\cos(D(B_{i_c}, M_i))|\cdot z_{i_c}$.
\item For $c\in\{1, 2, ..., C\}$, we calculate the difference in confidence score from each first-stage detector, $z'_i - z_{i_c}$, and concatenate those numbers.
\end{enumerate}

\ifCLASSOPTIONcompsoc
  \section*{Acknowledgments}
\else
  \section*{Acknowledgment}
\fi

This research is supported in part by the Office of Naval Research (ONR) under Award Numbers N00014-13-1-0644 and N00014-17-1-2094, and by a pilot grant from the Brown University Center for Vision Research.

\ifCLASSOPTIONcaptionsoff
  \newpage
\fi

\bibliographystyle{IEEEtran}
\bibliography{cog_ref}

\begin{thebibliography}{10}
\providecommand{\url}[1]{#1}
\csname url@samestyle\endcsname
\providecommand{\newblock}{\relax}
\providecommand{\bibinfo}[2]{#2}
\providecommand{\BIBentrySTDinterwordspacing}{\spaceskip=0pt\relax}
\providecommand{\BIBentryALTinterwordstretchfactor}{4}
\providecommand{\BIBentryALTinterwordspacing}{\spaceskip=\fontdimen2\font plus
\BIBentryALTinterwordstretchfactor\fontdimen3\font minus
  \fontdimen4\font\relax}
\providecommand{\BIBforeignlanguage}[2]{{%
\expandafter\ifx\csname l@#1\endcsname\relax
\typeout{** WARNING: IEEEtran.bst: No hyphenation pattern has been}%
\typeout{** loaded for the language `#1'. Using the pattern for}%
\typeout{** the default language instead.}%
\else
\language=\csname l@#1\endcsname
\fi
#2}}
\providecommand{\BIBdecl}{\relax}
\BIBdecl

\bibitem{pascal-voc-2012}
M.~Everingham, L.~Van~Gool, C.~K. Williams, J.~Winn, and A.~Zisserman, ``The
  pascal visual object classes (voc) challenge,'' \emph{International Journal
  of Computer Vision (IJCV)}, vol.~88, no.~2, pp. 303--338, 2010.

\bibitem{ILSVRC15}
O.~Russakovsky, J.~Deng, H.~Su, J.~Krause, S.~Satheesh, S.~Ma, Z.~Huang,
  A.~Karpathy, A.~Khosla, M.~Bernstein, A.~C. Berg, and L.~Fei-Fei, ``{ImageNet
  Large Scale Visual Recognition Challenge},'' \emph{International Journal of
  Computer Vision (IJCV)}, 2015.

\bibitem{lee2009geometric}
D.~C. Lee, M.~Hebert, and T.~Kanade, ``Geometric reasoning for single image
  structure recovery,'' in \emph{Proceedings of the IEEE Conference on Computer
  Vision and Pattern Recognition (CVPR)}.\hskip 1em plus 0.5em minus
  0.4em\relax IEEE, 2009, pp. 2136--2143.

\bibitem{hedau2010thinking}
V.~Hedau, D.~Hoiem, and D.~Forsyth, ``Thinking inside the box: Using appearance
  models and context based on room geometry,'' in \emph{Proceedings of the
  European Conference on Computer Vision (ECCV)}.\hskip 1em plus 0.5em minus
  0.4em\relax Springer, 2010, pp. 224--237.

\bibitem{zhang2013estimating}
J.~Zhang, C.~Kan, A.~G. Schwing, and R.~Urtasun, ``Estimating the {3D} layout
  of indoor scenes and its clutter from depth sensors,'' in \emph{Proceedings
  of the IEEE International Conference on Computer Vision (ICCV)}.\hskip 1em
  plus 0.5em minus 0.4em\relax IEEE, 2013, pp. 1273--1280.

\bibitem{silberman2012indoor}
N.~Silberman, D.~Hoiem, P.~Kohli, and R.~Fergus, ``Indoor segmentation and
  support inference from {RGBD} images,'' in \emph{Proceedings of the European
  Conference on Computer Vision (ECCV)}.\hskip 1em plus 0.5em minus 0.4em\relax
  Springer, 2012, pp. 746--760.

\bibitem{fouhey2014unfolding}
D.~F. Fouhey, A.~Gupta, and M.~Hebert, ``Unfolding an indoor origami world,''
  in \emph{Proceedings of the European Conference on Computer Vision
  (ECCV)}.\hskip 1em plus 0.5em minus 0.4em\relax Springer, 2014, pp. 687--702.

\bibitem{guo2013support}
R.~Guo and D.~Hoiem, ``Support surface prediction in indoor scenes,'' in
  \emph{Proceedings of the IEEE International Conference on Computer Vision
  (ICCV)}.\hskip 1em plus 0.5em minus 0.4em\relax IEEE, 2013, pp. 2144--2151.

\bibitem{gupta2014learning}
S.~Gupta, R.~Girshick, P.~Arbel{\'a}ez, and J.~Malik, ``Learning rich features
  from {RGB-D} images for object detection and segmentation,'' in
  \emph{Proceedings of the European Conference on Computer Vision
  (ECCV)}.\hskip 1em plus 0.5em minus 0.4em\relax Springer, 2014, pp. 345--360.

\bibitem{song2014sliding}
S.~Song and J.~Xiao, ``Sliding shapes for {3D} object detection in depth
  images,'' in \emph{Proceedings of the European Conference on Computer Vision
  (ECCV)}.\hskip 1em plus 0.5em minus 0.4em\relax Springer, 2014, pp. 634--651.

\bibitem{song2016ssc}
S.~Song, F.~Yu, A.~Zeng, A.~X. Chang, M.~Savva, and T.~Funkhouser, ``Semantic
  scene completion from a single depth image,'' \emph{Proceedings of the IEEE
  Conference on Computer Vision and Pattern Recognition (CVPR)}, 2017.

\bibitem{ren2016cog}
Z.~Ren and E.~B. Sudderth, ``Three-dimensional object detection and layout
  prediction using clouds of oriented gradients,'' in \emph{Proceedings of the
  IEEE Conference on Computer Vision and Pattern Recognition (CVPR)}, 2016, pp.
  1525--1533.

\bibitem{russell2009}
B.~C. Russell and A.~Torralba, ``Building a database of {3D} scenes from user
  annotations,'' in \emph{Proceedings of the IEEE Conference on Computer Vision
  and Pattern Recognition (CVPR)}.\hskip 1em plus 0.5em minus 0.4em\relax IEEE,
  2009, pp. 2711--2718.

\bibitem{lai2011}
K.~Lai, L.~Bo, X.~Ren, and D.~Fox, ``A large-scale hierarchical multi-view
  {RGB-D} object dataset,'' in \emph{IEEE International Conference on Robotics
  and Automation (ICRA)}.\hskip 1em plus 0.5em minus 0.4em\relax IEEE, 2011,
  pp. 1817--1824.

\bibitem{song2015sun}
S.~Song, L.~Samuel, and J.~Xiao, ``{SUN RGB-D}: A {RGB-D} scene understanding
  benchmark suite,'' in \emph{Proceedings of the IEEE Conference on Computer
  Vision and Pattern Recognition (CVPR)}.\hskip 1em plus 0.5em minus
  0.4em\relax IEEE, 2015.

\bibitem{Geiger2012kitti}
A.~Geiger, P.~Lenz, and R.~Urtasun, ``Are we ready for autonomous driving? the
  kitti vision benchmark suite,'' in \emph{Proceedings of the IEEE Conference
  on Computer Vision and Pattern Recognition (CVPR)}.\hskip 1em plus 0.5em
  minus 0.4em\relax IEEE, 2012, pp. 3354--3361.

\bibitem{song2016deep}
S.~Song and J.~Xiao, ``Deep sliding shapes for amodal {3D} object detection in
  {RGB-D} images,'' in \emph{Proceedings of the IEEE Conference on Computer
  Vision and Pattern Recognition (CVPR)}, 2016.

\bibitem{fidler20123d}
S.~Fidler, S.~Dickinson, and R.~Urtasun, ``{3D} object detection and viewpoint
  estimation with a deformable {3D} cuboid model,'' in \emph{Advances in Neural
  Information Processing Systems (NeurIPS)}, 2012, pp. 611--619.

\bibitem{Geiger2015GCPR}
A.~Geiger and C.~Wang, ``Joint {3D} object and layout inference from a single
  {RGB-D} image,'' in \emph{German Conference on Pattern Recognition (GCPR)},
  2015.

\bibitem{heitz2009cascaded}
G.~Heitz, S.~Gould, A.~Saxena, and D.~Koller, ``Cascaded classification models:
  Combining models for holistic scene understanding,'' in \emph{Advances in
  Neural Information Processing Systems (NeurIPS)}, 2009, pp. 641--648.

\bibitem{coughlan1999manhattan}
J.~M. Coughlan and A.~L. Yuille, ``Manhattan world: Compass direction from a
  single image by {B}ayesian inference,'' in \emph{Proceedings of the IEEE
  International Conference on Computer Vision (ICCV)}, vol.~2.\hskip 1em plus
  0.5em minus 0.4em\relax IEEE, 1999, pp. 941--947.

\bibitem{schwing2012efficient}
A.~G. Schwing, T.~Hazan, M.~Pollefeys, and R.~Urtasun, ``Efficient structured
  prediction for {3D} indoor scene understanding,'' in \emph{Proceedings of the
  IEEE Conference on Computer Vision and Pattern Recognition (CVPR)}.\hskip 1em
  plus 0.5em minus 0.4em\relax IEEE, 2012, pp. 2815--2822.

\bibitem{bai2012fast}
J.~Bai, Q.~Song, O.~Veksler, and X.~Wu, ``Fast dynamic programming for labeling
  problems with ordering constraints,'' in \emph{Proceedings of the IEEE
  Conference on Computer Vision and Pattern Recognition (CVPR)}.\hskip 1em plus
  0.5em minus 0.4em\relax IEEE, 2012, pp. 1728--1735.

\bibitem{dalal2005histograms}
N.~Dalal and B.~Triggs, ``Histograms of oriented gradients for human
  detection,'' in \emph{Proceedings of the IEEE Conference on Computer Vision
  and Pattern Recognition (CVPR)}, vol.~1.\hskip 1em plus 0.5em minus
  0.4em\relax IEEE, 2005, pp. 886--893.

\bibitem{felzenszwalb2010dpm}
P.~F. Felzenszwalb, R.~B. Girshick, D.~McAllester, and D.~Ramanan, ``Object
  detection with discriminatively trained part-based models,'' \emph{IEEE
  Transactions on Pattern Analysis and Machine Intelligence (TPAMI)}, vol.~32,
  no.~9, pp. 1627--1645, 2010.

\bibitem{girshick2014rich}
R.~Girshick, J.~Donahue, T.~Darrell, and J.~Malik, ``Rich feature hierarchies
  for accurate object detection and semantic segmentation,'' in
  \emph{Proceedings of the IEEE Conference on Computer Vision and Pattern
  Recognition (CVPR)}, 2014, pp. 580--587.

\bibitem{girshick15fastrcnn}
R.~Girshick, ``Fast {R-CNN},'' in \emph{Proceedings of the IEEE International
  Conference on Computer Vision (ICCV)}, 2015.

\bibitem{ren2015faster}
S.~Ren, K.~He, R.~Girshick, and J.~Sun, ``Faster {R-CNN}: Towards real-time
  object detection with region proposal networks,'' in \emph{Advances in Neural
  Information Processing Systems (NeurIPS)}, 2015.

\bibitem{he2016res}
K.~He, X.~Zhang, S.~Ren, and J.~Sun, ``Deep residual learning for image
  recognition,'' in \emph{Proceedings of the IEEE Conference on Computer Vision
  and Pattern Recognition (CVPR)}, 2016, pp. 770--778.

\bibitem{lin2017focal}
T.-Y. Lin, P.~Goyal, R.~Girshick, K.~He, and P.~Doll{\'a}r, ``Focal loss for
  dense object detection,'' in \emph{Proceedings of the IEEE International
  Conference on Computer Vision (ICCV)}, 2017.

\bibitem{redmon2016you}
J.~Redmon, S.~Divvala, R.~Girshick, and A.~Farhadi, ``You only look once:
  Unified, real-time object detection,'' in \emph{Proceedings of the IEEE
  Conference on Computer Vision and Pattern Recognition (CVPR)}, 2016, pp.
  779--788.

\bibitem{redmon2016yolo9000}
J.~Redmon and A.~Farhadi, ``Yolo9000: Better, faster, stronger,''
  \emph{Proceedings of the IEEE Conference on Computer Vision and Pattern
  Recognition (CVPR)}, 2017.

\bibitem{wu20143d}
Z.~Wu, S.~Song, A.~Khosla, X.~Tang, and J.~Xiao, ``{3D} shapenets for {2.5D}
  object recognition and next-best-view prediction,'' \emph{Proceedings of the
  IEEE Conference on Computer Vision and Pattern Recognition (CVPR)}, 2015.

\bibitem{su15mvcnn}
H.~Su, S.~Maji, E.~Kalogerakis, and E.~G. Learned{-}Miller, ``Multi-view
  convolutional neural networks for {3D} shape recognition,'' in
  \emph{Proceedings of the IEEE International Conference on Computer Vision
  (ICCV)}, 2015.

\bibitem{qi2017pointnet}
C.~R. Qi, H.~Su, K.~Mo, and L.~J. Guibas, ``Pointnet: Deep learning on point
  sets for {3D} classification and segmentation,'' \emph{Proceedings of the
  IEEE Conference on Computer Vision and Pattern Recognition (CVPR)}, 2017.

\bibitem{qi2017pointnetpp}
C.~R. Qi, L.~Yi, H.~Su, and L.~J. Guibas, ``Pointnet++: Deep hierarchical
  feature learning on point sets in a metric space,'' \emph{Advances in Neural
  Information Processing Systems (NeurIPS)}, 2017.

\bibitem{hao2013}
H.~Jiang and J.~Xiao, ``A linear approach to matching cuboids in {RGBD}
  images,'' in \emph{Proceedings of the IEEE Conference on Computer Vision and
  Pattern Recognition (CVPR)}, 2013.

\bibitem{jia2013}
Z.~Jia, A.~Gallagher, A.~Saxena, and T.~Chen, ``{3D}-based reasoning with
  blocks, support, and stability,'' in \emph{Proceedings of the IEEE Conference
  on Computer Vision and Pattern Recognition (CVPR)}.\hskip 1em plus 0.5em
  minus 0.4em\relax IEEE, 2013, pp. 1--8.

\bibitem{xiao2012}
J.~Xiao, B.~C. Russell, and A.~Torralba, ``Localizing 3d cuboids in single-view
  images.'' in \emph{Advances in Neural Information Processing Systems
  (NeurIPS)}, 2012.

\bibitem{lee2017roomnet}
C.-Y. Lee, V.~Badrinarayanan, T.~Malisiewicz, and A.~Rabinovich, ``Roomnet:
  End-to-end room layout estimation,'' 2017.

\bibitem{schwing2013box}
A.~G. Schwing, S.~Fidler, M.~Pollefeys, and R.~Urtasun, ``Box in the box: Joint
  {3D} layout and object reasoning from single images,'' in \emph{Proceedings
  of the IEEE International Conference on Computer Vision (ICCV)}.\hskip 1em
  plus 0.5em minus 0.4em\relax IEEE, 2013, pp. 353--360.

\bibitem{shao2014imagining}
T.~Shao, A.~Monszpart, Y.~Zheng, B.~Koo, W.~Xu, K.~Zhou, and N.~J. Mitra,
  ``Imagining the unseen: Stability-based cuboid arrangements for scene
  understanding,'' \emph{ACM Transactions on Graphics (SIGGRAPH ASIA)},
  vol.~33, no.~6, 2014.

\bibitem{zhang2017context}
Y.~Zhang, M.~Bai, P.~Kohli, S.~Izadi, and J.~Xiao, ``Deepcontext:
  Context-encoding neural pathways for {3D} holistic scene understanding,'' in
  \emph{Proceedings of the IEEE International Conference on Computer Vision
  (ICCV)}, Oct 2017.

\bibitem{pamishapeTulsianiKCM15}
S.~Tulsiani, A.~Kar, J.~Carreira, and J.~Malik, ``Learning category-specific
  deformable {3D} models for object reconstruction,'' \emph{IEEE Transactions
  on Pattern Analysis and Machine Intelligence (TPAMI)}, 2016.

\bibitem{dai2017scannet}
A.~Dai, A.~X. Chang, M.~Savva, M.~Halber, T.~Funkhouser, and M.~Nie{\ss}ner,
  ``Scannet: Richly-annotated {3D} reconstructions of indoor scenes,'' in
  \emph{Proceedings of the IEEE Conference on Computer Vision and Pattern
  Recognition (CVPR)}, 2017.

\bibitem{chen173dop}
X.~Chen, K.~Kundu, Y.~Zhu, H.~Ma, S.~Fidler, and R.~Urtasun, ``{3D} object
  proposals using stereo imagery for accurate object class detection,'' in
  \emph{IEEE Transactions on Pattern Analysis and Machine Intelligence
  (TPAMI)}, 2017.

\bibitem{mousavian20173d}
A.~Mousavian, D.~Anguelov, J.~Flynn, and J.~Ko{\v{s}}eck{\'a}, ``{3D} bounding
  box estimation using deep learning and geometry,'' in \emph{Proceedings of
  the IEEE Conference on Computer Vision and Pattern Recognition (CVPR)}.\hskip
  1em plus 0.5em minus 0.4em\relax IEEE, 2017, pp. 5632--5640.

\bibitem{xiang2015data}
Y.~Xiang, W.~Choi, Y.~Lin, and S.~Savarese, ``Data-driven {3D} voxel patterns
  for object category recognition,'' in \emph{Proceedings of the IEEE
  Conference on Computer Vision and Pattern Recognition (CVPR)}, 2015, pp.
  1903--1911.

\bibitem{qi2017frustum}
C.~R. Qi, W.~Liu, C.~Wu, H.~Su, and L.~J. Guibas, ``Frustum pointnets for 3d
  object detection from rgb-d data,'' 2018.

\bibitem{zhou2017voxelnet}
Y.~Zhou and O.~Tuzel, ``Voxelnet: End-to-end learning for point cloud based 3d
  object detection,'' \emph{Proceedings of the IEEE Conference on Computer
  Vision and Pattern Recognition (CVPR)}, 2018.

\bibitem{chen17multiview3D}
X.~Chen, H.~Ma, J.~Wan, B.~Li, and T.~Xia, ``Multi-view {3D} object detection
  network for autonomous driving,'' in \emph{Proceedings of the IEEE Conference
  on Computer Vision and Pattern Recognition (CVPR)}, 2017.

\bibitem{lin2013holistic}
D.~Lin, S.~Fidler, and R.~Urtasun, ``Holistic scene understanding for {3D}
  object detection with {RGBD} cameras,'' in \emph{Proceedings of the IEEE
  International Conference on Computer Vision (ICCV)}.\hskip 1em plus 0.5em
  minus 0.4em\relax IEEE, 2013, pp. 1417--1424.

\bibitem{gupta2010blocks}
A.~Gupta, A.~A. Efros, and M.~Hebert, ``Blocks world revisited: Image
  understanding using qualitative geometry and mechanics,'' in
  \emph{Proceedings of the European Conference on Computer Vision
  (ECCV)}.\hskip 1em plus 0.5em minus 0.4em\relax Springer, 2010, pp. 482--496.

\bibitem{guptaCVPR15align}
S.~Gupta, P.~A. Arbel{\'{a}}ez, R.~B. Girshick, and J.~Malik, ``Aligning {3D}
  models to {RGB-D} images of cluttered scenes,'' in \emph{Proceedings of the
  IEEE Conference on Computer Vision and Pattern Recognition (CVPR)}, 2015.

\bibitem{Aubry14chair}
M.~Aubry, D.~Maturana, A.~Efros, B.~Russell, and J.~Sivic, ``Seeing {3D}
  chairs: Exemplar part-based {2D-3D} alignment using a large dataset of {CAD}
  models,'' in \emph{Proceedings of the IEEE Conference on Computer Vision and
  Pattern Recognition (CVPR)}, 2014.

\bibitem{lim2013parsing}
J.~J. Lim, H.~Pirsiavash, and A.~Torralba, ``Parsing {IKEA} objects: Fine pose
  estimation,'' in \emph{Proceedings of the IEEE International Conference on
  Computer Vision (ICCV)}, 2013.

\bibitem{lim2014fpm}
J.~J. Lim, A.~Khosla, and A.~Torralba, ``{FPM}: Fine pose parts-based model
  with {3D CAD} models,'' in \emph{Proceedings of the European Conference on
  Computer Vision (ECCV)}.\hskip 1em plus 0.5em minus 0.4em\relax Springer,
  2014, pp. 478--493.

\bibitem{maturana2015voxnet}
D.~Maturana and S.~Scherer, ``Voxnet: A {3D} convolutional neural network for
  real-time object recognition,'' in \emph{IEEE/RSJ International Conference on
  Intelligent Robots and Systems}.\hskip 1em plus 0.5em minus 0.4em\relax IEEE,
  2015, pp. 922--928.

\bibitem{zhuo17amodal3d}
Z.~Deng and L.~J. Latecki, ``Amodal detection of {3D} objects: Inferring {3D}
  bounding boxes from 2d ones in rgb-depth images,'' in \emph{Proceedings of
  the IEEE Conference on Computer Vision and Pattern Recognition (CVPR)}, 2017.

\bibitem{lahoud20172d}
J.~Lahoud and B.~Ghanem, ``{2D}-driven {3D} object detection in {RGB}-{D}
  images,'' in \emph{Proceedings of the IEEE International Conference on
  Computer Vision (ICCV)}, Oct 2017.

\bibitem{wang2015designing}
X.~Wang, D.~Fouhey, and A.~Gupta, ``Designing deep networks for surface normal
  estimation,'' in \emph{Proceedings of the IEEE Conference on Computer Vision
  and Pattern Recognition (CVPR)}, 2015.

\bibitem{bansal2016marr}
A.~Bansal, B.~Russell, and A.~Gupta, ``Marr revisited: {2D}-{3D} alignment via
  surface normal prediction,'' in \emph{Proceedings of the IEEE Conference on
  Computer Vision and Pattern Recognition (CVPR)}, 2016, pp. 5965--5974.

\bibitem{hedau2009recovering}
V.~Hedau, D.~Hoiem, and D.~Forsyth, ``Recovering the spatial layout of
  cluttered rooms,'' in \emph{Proceedings of the IEEE Conference on Computer
  Vision and Pattern Recognition (CVPR)}.\hskip 1em plus 0.5em minus
  0.4em\relax IEEE, 2009, pp. 1849--1856.

\bibitem{mallya2015learning}
A.~Mallya and S.~Lazebnik, ``Learning informative edge maps for indoor scene
  layout prediction,'' in \emph{Proceedings of the IEEE International
  Conference on Computer Vision (ICCV)}, 2015, pp. 936--944.

\bibitem{zou2018layout}
C.~Zou, A.~Colburn, Q.~Shan, and D.~Hoiem, ``Layoutnet: Reconstructing the 3d
  room layout from a single rgb image,'' 2018.

\bibitem{yao2012describing}
J.~Yao, S.~Fidler, and R.~Urtasun, ``Describing the scene as a whole: Joint
  object detection, scene classification and semantic segmentation,'' in
  \emph{Proceedings of the IEEE Conference on Computer Vision and Pattern
  Recognition (CVPR)}.\hskip 1em plus 0.5em minus 0.4em\relax IEEE, 2012, pp.
  702--709.

\bibitem{hoiem2008putting}
D.~Hoiem, A.~A. Efros, and M.~Hebert, ``Putting objects in perspective,''
  \emph{International Journal of Computer Vision (IJCV)}, vol.~80, no.~1, pp.
  3--15, 2008.

\bibitem{yuzhuo2018C3D}
Y.~Ren, C.~Chen, S.~Li, and C.-C.~J. Kuo, ``Context-assisted 3d (c3d) object
  detection from rgb-d images,'' \emph{Journal of Visual Communication and
  Image Representation}, vol.~34, no.~11, pp. 2189--2202, 2012.

\bibitem{johnson1999using}
A.~E. Johnson and M.~Hebert, ``Using spin images for efficient object
  recognition in cluttered {3D} scenes,'' \emph{IEEE Transactions on Pattern
  Analysis and Machine Intelligence (TPAMI)}, vol.~21, no.~5, pp. 433--449,
  1999.

\bibitem{payet2011contours}
N.~Payet and S.~Todorovic, ``From contours to 3d object detection and pose
  estimation,'' in \emph{Proceedings of the IEEE International Conference on
  Computer Vision (ICCV)}.\hskip 1em plus 0.5em minus 0.4em\relax IEEE, 2011,
  pp. 983--990.

\bibitem{buch20093d}
N.~Buch, J.~Orwell, and S.~A. Velastin, ``{3D} extended histogram of oriented
  gradients (3dhog) for classification of road users in urban scenes,'' in
  \emph{Proceedings of the British Machine Vision Conference (BMVC)}, 2009.

\bibitem{scherer2010histograms}
M.~Scherer, M.~Walter, and T.~Schreck, ``Histograms of oriented gradients for
  {3D} object retrieval,'' in \emph{Proceedings of the European Conference on
  Computer Vision (ECCV)}, 2010.

\bibitem{song2016}
S.~Song and J.~Xiao, ``Deep sliding shapes for amodal {3D} object detection in
  {RGB-D} images,'' in \emph{Proceedings of the IEEE Conference on Computer
  Vision and Pattern Recognition (CVPR)}, 2016.

\bibitem{alexe2012measuring}
B.~Alexe, T.~Deselaers, and V.~Ferrari, ``Measuring the objectness of image
  windows,'' \emph{IEEE Transactions on Pattern Analysis and Machine
  Intelligence (TPAMI)}, vol.~34, no.~11, pp. 2189--2202, 2012.

\bibitem{kuo2015deepbox}
W.~Kuo, B.~Hariharan, and J.~Malik, ``Deepbox: Learning objectness with
  convolutional networks,'' in \emph{Proceedings of the IEEE International
  Conference on Computer Vision (ICCV)}, 2015, pp. 2479--2487.

\bibitem{mostajabi2015feedforward}
M.~Mostajabi, P.~Yadollahpour, and G.~Shakhnarovich, ``Feedforward semantic
  segmentation with zoom-out features,'' in \emph{Proceedings of the IEEE
  Conference on Computer Vision and Pattern Recognition (CVPR)}, 2015, pp.
  3376--3385.

\bibitem{joachims2009cutting}
T.~Joachims, T.~Finley, and C.-N.~J. Yu, ``Cutting-plane training of structural
  svms,'' \emph{Machine Learning}, 2009.

\bibitem{vedaldi09structured}
A.~Vedaldi and A.~Zisserman, ``Structured output regression for detection with
  partial occulsion,'' in \emph{Advances in Neural Information Processing
  Systems (NeurIPS)}, 2009.

\bibitem{yu2009learning}
C.-N.~J. Yu and T.~Joachims, ``Learning structural svms with latent
  variables,'' in \emph{International Conference on Machine Learning
  (ICML)}.\hskip 1em plus 0.5em minus 0.4em\relax ACM, 2009, pp. 1169--1176.

\bibitem{yuille2003concave}
A.~Yuille and A.~Rangarajan, ``The concave-convex procedure,'' \emph{Neural
  Computation}, vol.~15, no.~4, pp. 915--936, 2003.

\bibitem{chen2016r}
C.~Chen, M.-Y. Liu, O.~Tuzel, and J.~Xiao, ``{R-CNN} for small object
  detection,'' in \emph{ACCV}.\hskip 1em plus 0.5em minus 0.4em\relax Springer,
  2016, pp. 214--230.

\bibitem{hu2016finding}
P.~Hu and D.~Ramanan, ``Finding tiny faces,'' \emph{Proceedings of the IEEE
  Conference on Computer Vision and Pattern Recognition (CVPR)}, 2017.

\bibitem{pero2011sampling}
L.~D. Pero, J.~Guan, E.~Brau, J.~Schlecht, and K.~Barnard, ``Sampling
  bedrooms,'' in \emph{Proceedings of the IEEE Conference on Computer Vision
  and Pattern Recognition (CVPR)}.\hskip 1em plus 0.5em minus 0.4em\relax IEEE,
  2011, pp. 2009--2016.

\bibitem{nowozin2011structured}
S.~Nowozin and C.~H. Lampert, ``Structured learning and prediction in computer
  vision,'' \emph{Foundations and Trends in Computer Graphics and Vision},
  vol.~6, no. 3--4, pp. 185--365, 2011.

\bibitem{rabinovich2007objects}
A.~Rabinovich, A.~Vedaldi, C.~Galleguillos, E.~Wiewiora, and S.~Belongie,
  ``Objects in context,'' in \emph{Proceedings of the IEEE International
  Conference on Computer Vision (ICCV)}, 2007.

\bibitem{ren2018latent}
Z.~Ren and E.~B. Sudderth, ``{3D} object detection with latent support
  surfaces,'' in \emph{Proceedings of the IEEE Conference on Computer Vision
  and Pattern Recognition (CVPR)}, 2018.

\end{thebibliography}

\end{document}